\pdfoutput=1
\documentclass[11pt]{article}

\usepackage{EMNLP2023}
\usepackage{times}
\usepackage{latexsym}

\usepackage[T1]{fontenc}
\usepackage[utf8]{inputenc}

\usepackage{microtype}

\usepackage{inconsolata}
\usepackage{comment}
\usepackage{xspace}
\usepackage{multirow}
\usepackage{multicol}
\usepackage{graphicx}
\usepackage{booktabs}
\usepackage{todonotes}
\usepackage{tablefootnote}
\usepackage{soul}
\usepackage{subfigure}
\usepackage{listings}
\usepackage{pgfplots}

\usepackage{listings}
\definecolor{codegreen}{rgb}{0,0.6,0}
\definecolor{codegray}{rgb}{0.5,0.5,0.5}
\definecolor{codepurple}{rgb}{0.58,0,0.82}

\lstdefinestyle{mystyle}{
  frame=single,
  basicstyle=\ttfamily\footnotesize,
  backgroundcolor=\color{backcolour}, commentstyle=\color{codegreen},
  commentstyle=\color{darkgreen}\slshape,
  keywordstyle=\color{blue},
  stringstyle=\color{darkred},
  numberstyle=\tiny\color{codegray},
  emphstyle=\color{pink}\underbar,
  morekeywords={Verify, Question},
  escapeinside={(*@}{@*)},
  breakatwhitespace=false,         
  breaklines=true,                 
  captionpos=b,                    
  keepspaces=true,                    
  numbersep=5pt,                  
  showspaces=false,                
  showstringspaces=false,
  showtabs=false,                  
  tabsize=2
}
\newcommand{\authorspace}{\hspace{0.3cm}}
\newcommand*\samethanks[1][\value{footnote}]{\footnotemark[#1]}
\newcommand{\dataset}{\textsc{SciTab}\xspace}
\definecolor{darkgreen}{rgb}{0,0.50,0}
\definecolor{kmy-color}{rgb}{0.858, 0.188, 0.478}

\title{\dataset: A Challenging Benchmark for Compositional Reasoning\\ and Claim Verification on Scientific Tables}

\author{
    \bf Xinyuan Lu\thanks{~~Equal Contribution.}$\:\,^{1,2}$\authorspace
  \bf Liangming Pan\samethanks$\:\,^3$ \authorspace
  \bf Qian Liu$^4$ \authorspace
    \\ \vspace{2mm}
  \bf Preslav Nakov$^5$ \authorspace
  \bf Min-Yen Kan$^2$ \authorspace
  \\
 $^1$ISEP Program, NUS Graduate School \authorspace
 $^2$ National University of Singapore \\
 $^3$University of California, Santa Barbara \authorspace 
 $^4$Sea AI Lab \authorspace
 $^5$MBZUAI 
 \vspace{2mm}
 \\
  {\tt luxinyuan@u.nus.edu} \authorspace 
  {\tt liangmingpan@ucsb.edu} \authorspace
 {\tt liuqian@sea.com}  \\
  {\tt preslav.nakov@mbzuai.ac.ae} \authorspace
  {\tt kanmy@comp.nus.edu.sg}
}

\begin{document}
\maketitle

\begin{abstract}
Current scientific fact-checking benchmarks exhibit several shortcomings, such as biases arising from crowd-sourced claims and an over-reliance on text-based evidence. We present \dataset, a challenging evaluation dataset consisting of 1.2K expert-verified scientific claims that 1) originate from authentic scientific publications and 2) require compositional reasoning for verification. The claims are paired with evidence-containing scientific tables annotated with labels. Through extensive evaluations, we demonstrate that \dataset poses a significant challenge to state-of-the-art models, including table-based pretraining models and large language models. All models except GPT-4 achieved performance barely above random guessing. Popular prompting techniques, such as Chain-of-Thought, do not achieve much performance gains on \dataset. Our analysis uncovers several unique challenges posed by \dataset, including table grounding, claim ambiguity, and compositional reasoning. Our codes and data are publicly available at \url{https://github.com/XinyuanLu00/SciTab}.

% By making \dataset available to the research community, we aim to drive progress in the field of scientific fact-checking and enable the development of more sophisticated and capable automated fact-checking systems.
% In this paper, we introduce \dataset, a novel dataset that addresses these issues by incorporating real-world claims from scientific papers along with quantitative data presented in tables. 

% [v3]
% Scientific fact-checking is crucial for ensuring the accuracy, reliability, and trustworthiness of scientific claims.  However, existing benchmarks are limited in terms of their claim diversity, reliance on text-based evidence, and oversimplification of scientific reasoning. To address these gaps, we introduce \dataset, a novel dataset comprising 1,225 challenging scientific claims requiring compositional reasoning with scientific tables. The claims in \dataset are derived from actual scientific statements, and the evidence is presented as scientific tables, closely mirroring real-world fact-checking scenarios. We establish benchmarks on \dataset using state-of-the-art models, revealing their inherent difficulty and highlighting limitations in existing prompting methods. Our error analysis identifies unique challenges, including ambiguous expressions and irrelevant claims, suggesting future research directions.
\end{abstract}

\section{Introduction}

% What is Sci-Fact-Check and why it is important?

Scientific fact-checking is a crucial process that involves validating the accuracy of scientific claims by cross-referencing them with established scientific literature, research, or data~\cite{DBLP:journals/tacl/GuoSV22}. This process is crucial for preserving the integrity of scientific information, preventing the spread of misinformation, and fostering public trust in research findings. However, the sheer volume of scientific data and claims can be overwhelming for manual fact-checking, making automated scientific fact-checking an imperative research area of NLP. 

%Context
% Scientific fact-checking assists users in evaluating the veracity of scientific claims by comparing them with relevant research literature~\cite{DBLP:journals/tacl/GuoSV22,DBLP:conf/acl/PanCXKW20}. Scientific claims are intrinsically linked to experimental data, which is often represented in tables and figures. By automatically verifying whether scientific claims are grounded in the tables and figures from which they originate, we can increase the reliability of research findings and strengthen public trust in research outcomes. 

%Review+Gap
Scientific fact-checking has advanced significantly with benchmarks including Sci-Fact~\cite{DBLP:conf/emnlp/WaddenLLWZCH20}, Sci-Fact Open~\cite{DBLP:conf/emnlp/WaddenLKCBWH22}, and COVID-Fact~\cite{DBLP:conf/acl/SaakyanCM20}. However, these datasets still exhibit several limitations. First, the claims are \textit{crowd-sourced} rather than collected from real scientific papers. This leads to problems such as bias in human annotation, a lack of diversity, and shallow claims that do not reflect the complexity of scientific reasoning.
For example, most claims in Sci-Fact can be validated by a single sentence in a paper's abstract, which oversimplifies the scientific discourse. Second, the claims in the existing benchmarks are solely validated against \textit{text-based evidence}, primarily paper abstracts. However, in many scientific processes, claims are intrinsically tied to quantitative experimental data, commonly presented in tables and figures. This disparity highlights a significant gap between the existing benchmarks and real-world scientific fact-checking needs. To bridge these gaps, a dataset that 1) compiles real-world claims from scientific papers, and 2) includes original scientific data such as tables and figures, is needed. 
% \qian{Maybe it will be better to present an example in the top right in this page.}

% these benchmarks validate claims using only \textit{text-based evidence}, mostly paper abstracts. 
% This reveals a significant gap between these benchmarks and real-world scientific fact-checking needs. 
% it is crucial to create a dataset that: 1) compiles real-world claims from scientific papers, and 2) takes into account original data, such as tables and figures, which form the basis of scientific claims. 

\begin{figure*}[!t]
    \centering
    \includegraphics[width=16cm]{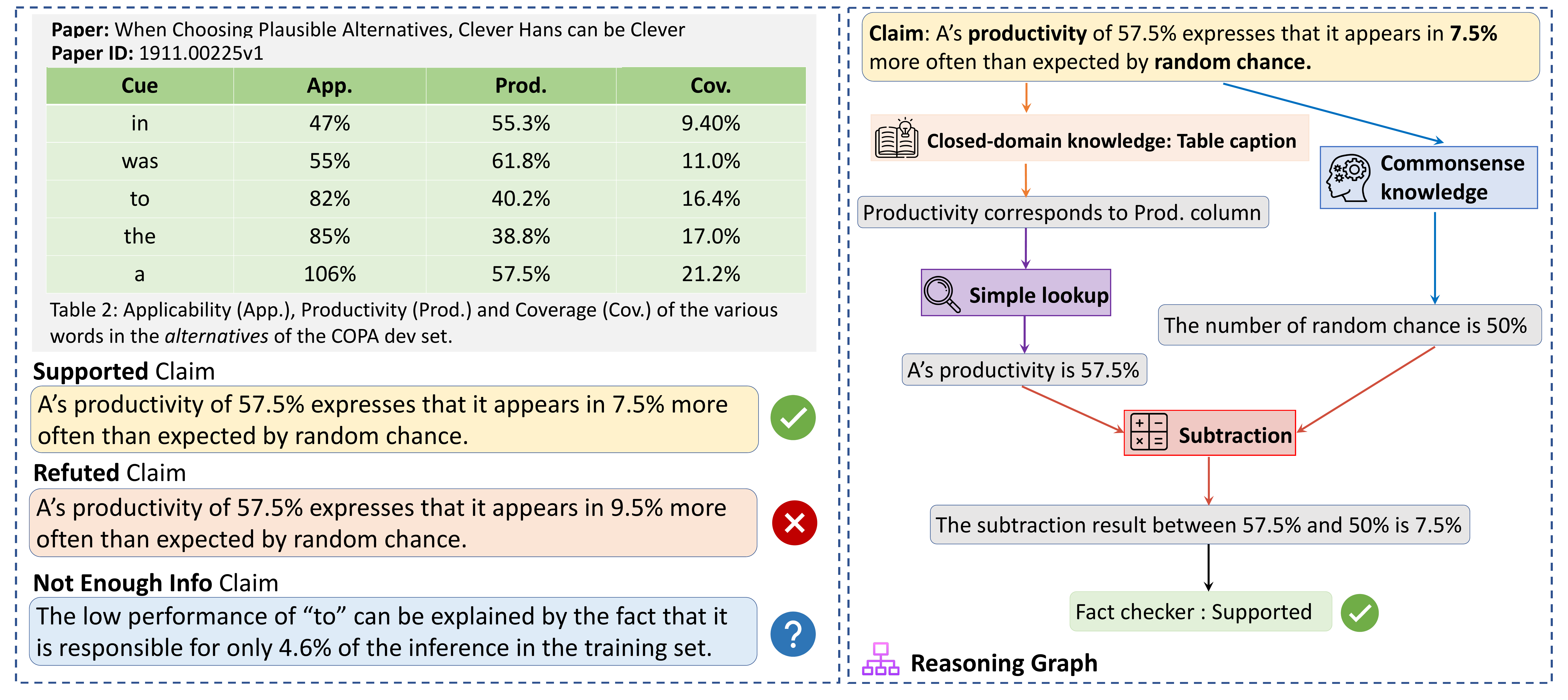}
    \caption{An example of our \dataset dataset (left) and its corresponding reasoning graph (right). Each data entry contains \textit{paper name}, \textit{paper id}, 
 \textit{table}, one \textit{claim}, and its corresponding \textit{label} (Supported, Refuted, Not Enough Info). 
    %\qian{It may be better to expand the term NEI here and in the figure.}
    }
    \label{fig:data_example}
\end{figure*}

\begin{figure*}[!thb]
    \centering
    \includegraphics[width=15cm]{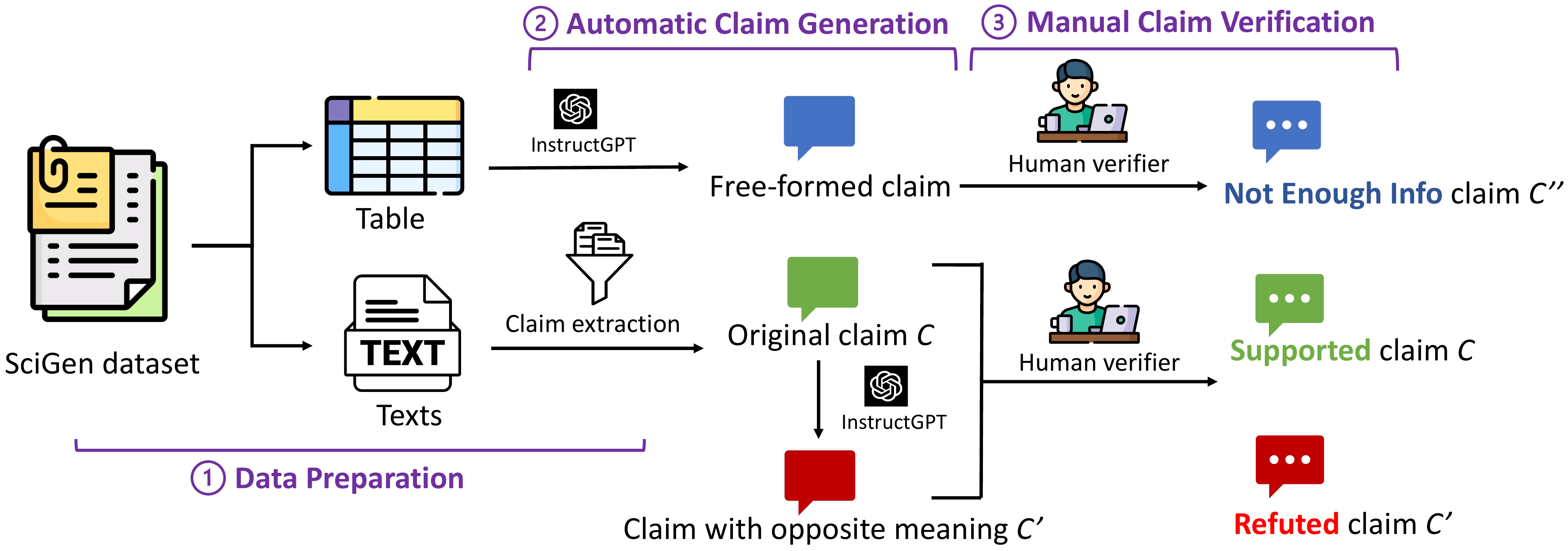}
    \caption{The \textit{human--model collaboration} construction process of \dataset, which contains three steps: 1) data preparation (including data preprocessing and claim extraction) 2) automatic claim generation (including refuted and Not Enough Info claim generation) and 3) manual claim verification.}
    \label{fig:annotation pipeline}
\end{figure*}

In this paper, we propose a novel dataset \dataset, which fulfills these stated criteria. It contains 1,225 challenging scientific claims, each demanding compositional reasoning for verification using scientific tables. Our data is derived from the SciGen dataset~\cite{DBLP:conf/nips/MoosaviRRG21}, a resource that includes scientific tables and claims crawled from arXiv.org. We first manually filter out the check-worthy scientific claims from the raw data. Following this, we employ a strategy of 
% Min: NN compound, use endash
human--model collaboration, as depicted in Figure~\ref{fig:annotation pipeline}, to generate claims that are either contradicted or unverifiable based on the table's content. Figure~\ref{fig:data_example} shows a claim from \dataset and the corresponding reasoning process to verify it. Compared with existing benchmarks, \dataset is closer to real-world scientific fact-checking in terms of more realistic claims and table-based evidence. Through data analysis, we further show that the claims in \dataset necessitate a more comprehensive and nuanced set of reasoning skills for verification, \textit{e.g.,} numerical reasoning and commonsense knowledge, \textit{etc}.
% The claims are constructed from actual scientific statements authored by researchers, while the evidence is presented in the form of scientific tables as opposed to abstracts. 

% With the $\sim$1.2K challenging, expert-annotated scientific claims, 
We employ \dataset as a diagnostic dataset for benchmarking the zero-shot and in-context learning performance for a wide range of state-of-the-art models, including table-based pretraining models, encoder–decoder models, open source language models, and API-based language models. 
% We establish benchmarks on our dataset for a wide range of state-of-the-art models, including table-based pretraining models, encoder--decoder models, decoder-only language models, and API-based language models. 
We observe that all models, with the exception of GPT-4, can only achieve marginally superior $F_1$ scores than random guessing, which underscores the challenging nature of~\dataset. Additionally, established prompting methods like Chain-of-Thought~\cite{DBLP:CoT} and Program-of-Thought~\cite{chen2022program} which typically enhance performance across most reasoning tasks, do not bring performance gain on~\dataset. Our error analysis sheds light on several unique challenges in \dataset that may lead to this, such as table grounding, dealing with ambiguous claims, and compositional reasoning. We make our dataset fully accessible to the research community. 

% In the spirit of advancing future research in scientific fact-checking, we make our dataset fully accessible to the research community.

%\kmy{When you evaluate a dataset, you are evaluating its quality. I think you mean you evaluate models on the dataset. Be precise.} 
% We also evaluated several state-of-the-art models on our dataset, including table-based models, open-source text-based models, and Large Language Models (LLMs). The results demonstrate that our dataset presents a challenging task for current LLMs, with challenges including writing approximation words to precise code, handling NEI claims, and solving entity linking issues.

% The main contributions of this work are two-fold:
% \begin{itemize}
% \item We introduce \textbf{\dataset}, a new dataset for complex scientific table fact verification, consisting of 1,225 samples. The dataset contains deep reasoning claims that make it more challenging than existing datasets in this domain, thereby posing a significant challenge for current LLMs.
% % \item We demonstrate that \dataset contains deep reasoning claims that make it more challenging than existing datasets in this domain.
% \item We conduct experiments with various baseline models on \dataset and demonstrate that they still lag behind the expert performance, highlighting the need for future research in this area.
% \end{itemize}

\section{The \dataset Dataset}

% Drop extraneous, zero-entropy phrases
% In this section, w
We adopt a \textit{human--model collaboration} strategy to construct \dataset, as shown in Figure~\ref{fig:annotation pipeline}. We describe the steps involved in data preparation (Section~\ref{sec:data preparation}), automatic claim generation (Section~\ref{sec:claim generation}), and manual claim verification (Section~\ref{sec:human verification}).

% In this section, we first describe how we crawl tables with corresponding claims and generate refuted and NEI scientific claims (Section~\ref{sec:data preparation}). Then, we illustrate the dataset annotation process (Section~\ref{sec:annotation process}). Last, the statistics of \dataset are described in Section~\ref{sec:data statistics}. The overall pipeline is shown in Figure~\ref{fig:annotation pipeline}.

\subsection{Data Preparation}
\label{sec:data preparation}
% \lxy{subheader remove; navigation paragraph shrink}
% \paragraph{Data Source.}
%\lxy{highlighting there are two parts of the dataset; Table + Claims; currently using `descriptions' but it is different from claims}
%describe SciGen
% \kmy{repetitive.  You already said this.  Bring in the detail from the Appendix and replace this text.}
% The \dataset is developed based on publicly available computer science papers collected in the SciGen dataset~\cite{DBLP:conf/nips/MoosaviRRG21}, which includes original tables and descriptions highlighted by the paper's authors. However, our dataset focuses only on claims worth verifying
% \lxy{what are the claims worth verifying, adding the definition}, which are different from the descriptions. To extract these claims based on the definitions in academic writing (see Appendix~\ref{append: claim definition}), we tokenized the descriptions into single sentences using the NLTK\footnote{\url{https://www.nltk.org/}} toolkit, resulting in 1,301 sentences.

% publicly available 
We use the publicly available SciGen~\cite{DBLP:conf/nips/MoosaviRRG21} dataset as our primary data source. The dataset was created by crawling computer science papers from arXiv. The tables and the texts explaining the tables are extracted from the papers to create (table, description) pairs for the task of data-to-text generation. From all the table descriptions of SciGen, we first filter the check-worthy scientific claims following the criteria established by~\citet{lee2009research} for academic writing\footnote{Detailed criteria are given in  Appendix~\ref{append: claim definition}}. We focus on the descriptions that serve the purpose of ``highlighting and commenting on key data'', \textit{i.e.}, describing research findings based on the data presented in scientific tables. 
% For the 1,301 table descriptions from SciGen, we hire a graduate student majoring in computer science to manually select scientific claims based on the aforementioned criteria using the user interface in Appendix~\ref{append:claim classification interface}. This gives us 872 real-world scientific claims together with the evidence tables. 
Given the task's objective nature and to save the cost of human labor, we hire a graduate student majoring in computer science to manually select scientific claims based on the aforementioned criteria using the user interface in Appendix~\ref{append:claim classification interface}. This decision was based on a pilot annotation which showed that a well-trained annotator can achieve over 95\% accuracy in filtering scientific claims. To safeguard the quality, we include an option to mark the claim as ``Discard-It's not a claim, or it's an incomplete, or not grammatically correct sentence.'' during the subsequent claim verification process. Using this approach, we filtered out 872 real-world scientific claims from 1,301 table descriptions in the SciGen dataset. 

\subsection{Automatic Claim Generation}
\label{sec:claim generation}
%describe STEP 2 and 3, generation prompts
% We consider those extracted claims to be Supported claims by default as they are from the papers themselves. \lxy{highlight the real-world refuted claims is very scarce: Due to the inherent characteristics of authentic academic papers, the identification of refuted or NEI claims poses significant challenges. Unless a claim contains a factual error, it is uncommon to encounter explicit instances of refutation in well-established scholarly publications. Consequently, the collection of naturally occurring refuted claims becomes arduous in practice.}
% To build a fact verification dataset, we also need to generate Refuted claims and Not Enough Info (NEI) claims. We now describe how we generate them.
%\lxy{New version: highlight the real-world refuted cases is seldom found, and prior works also generate refuted claims, so we need to generate them.}

\paragraph{False Claims.}
A fact-checking dataset requires both true and false claims. However, acquiring false claims that naturally occur within well-verified scientific publications is a challenging task. Following SciFact~\cite{DBLP:conf/emnlp/WaddenLLWZCH20} and COVID-Fact~\cite{DBLP:conf/acl/SaakyanCM20}, we seek to create false claims by generating counter-claims of the original true claims. Unlike previous works that purely rely on crowd-workers to compose counter-claims --- a process that is costly and prone to annotation artifacts --- we leverage the strong instruction-following capabilities of large language models (LLMs) to assist humans in generating candidate counter-claims. Specifically, we prompt \texttt{InstructGPT}~\cite{ouyang2022training} with the original claim and the instruction: \textit{Please modify the original claims to convey the opposite meaning with minimum edits}. To foster a varied set of generated claims, we include five diverse in-context examples and employ a high decoding temperature setting of $0.7$. By mandating minimal edits, we ensure that the counter-claims remain lexically close to the original claims, which is crucial in preventing fact-checking models from relying on superficial lexical patterns for verification.

\paragraph{Unverifiable Claims.}
To construct a more challenging dataset, we also integrate claims that are \textit{unverifiable} with the table information (labeled as Not Enough Info, \texttt{NEI}). We leverage \texttt{InstructGPT} to generate candidate NEI claims by prompting the model with the original table and the instruction: \textit{Please generate 5 relevant scientific claims based on the information in the table}. This process yields a diverse set of free-formed claims that enrich the diversity of \dataset. However, as LLMs tend to generate content that might not always be grounded in the provided data, many of the generated claims turn out to be \textit{relevant but unverifiable} with respect to the table. We adopt manual verification (elaborated in Section~\ref{sec:human verification}) to select them as NEI claims. 

% To expand the dataset with additional NEI claims, we utilize the InstructGPT (text-davinci-003) model, leveraging its text generation capabilities. By providing the original table evidence as input and prompting the model with the instruction to generate 10 claims based on the given table, we obtain 900 free-formed claims. These claims contribute to the diversity and richness of \dataset as they are derived from the hallucinations or speculative responses generated by the language model. This feature allows us to incorporate a wider range of perspectives and potential information that may not be explicitly stated in the original dataset.

% Similar to the validation process for refuted claims, we ensure that all the generated NEI claims adhere to correct grammar and semantics. 

\begin{table*}[!thb]
%\resizebox{0.5\textwidth}{!}{
\centering
\small
\begin{tabular}{llccccc}
\toprule
\multicolumn{2}{c}{\textbf{Statistics}}&\textbf{TabFact%~\cite{DBLP:conf/iclr/ChenWCZWLZW20}
} 
& \textbf{FEVEROUS%~\cite{DBLP:conf/nips/AlyGST00CM21}
}
&  \textbf{SEM-TAB-FACTS%~\cite{DBLP:conf/semeval/WangMDR21} 
} & \textbf{\textbf{\dataset}} \\
\midrule
\multicolumn{2}{l}{Domain} & Wiki Tables & Wiki Tables &  Scientific Articles &  Scientific Articles \\
\multicolumn{2}{l}{Annotator} & AMT & AMT & AMT & Experts \\
\multicolumn{2}{l}{Max. Reasoning Hops} &7  & 2 & 1 & 11\\
% Veracity Labels & & & & \\
\multirow{3}{*}
& Supported &  54\% & 56\% & 58\% & 37\%  \\ 
Veracity & Refuted & 46\%  & 39\%  & 38\%  & 34\%\\
& NEI & --- & 5\% & 4\%  & 29\% \\
%Total \# of Tables &  16,573 & 11.8M & 1,085 & 199\\
\multicolumn{2}{l}{Total \# of Claims}  &117,854 & 87,026& 5,715 & 1,225\\
\multicolumn{2}{l}{Avg. claims per table } & 7.11 & 0.07 & 5.27 & 6.16\\
\bottomrule
\end{tabular}
%}
\caption{Comparison of \dataset to three recent table fact verification datasets: TabFact \cite{DBLP:conf/iclr/ChenWCZWLZW20}, FEVEROUS \cite{DBLP:conf/nips/AlyGST00CM21}, and SEM-TAB-FACTS \cite{DBLP:conf/semeval/WangMDR21}. The table presents statistics related to the domain, annotator (AMT represents Amazon Mechanical Turk), maximum reasoning hops, veracity labels percentage of each dataset, the total number of claims, and average claims per table. }
\label{tab:dataset comparison}
\vspace{-0.3cm}
\end{table*}

\subsection{Manual Claim Verification}
\label{sec:human verification}

We subsequently employ a human verification process for two purposes: first, to verify the quality of the 872 false claims and 900 NEI claims that were generated by \texttt{InstructGPT}; second, to critically review the 872 real-world scientific claims obtained in Section~\ref{sec:data preparation}. This task involves selecting claims that can be verified exclusively based on the information presented in the table, without the need for additional context from the associated paper. 

For each pair of the true claim $c$ and its corresponding generated counter-claim $c'$, we ask the annotator to choose one of the following three options: (A) $c$ is not exclusively supported by the table, (B) $c$ is exclusively supported by the table, but $c'$ is not refuted by the table, and (C) $c$ is not exclusively supported by the table, and $c'$ is not refuted by the table. For each candidate NEI claim, we ask the annotator to judge whether it is unverifiable with respect to the table. 

% Given the 872 and 898 candidate false claims and NEI claims generated by \texttt{InstructGPT}, we employ a human verification process to ensure the accuracy and reliability of the generated claims. 

% Due to the relatively low number of claims available, we employed domain experts to manually verify the generated refuted and NEI claims. 

\paragraph{Annotator Recruitment.} 
Given that our data source is from computer science papers, we recruit university students majoring in computer science with basic math and programming backgrounds for annotation. We ask each annotator to fill in a questionnaire, including their age, department, maximum workload per week, etc. 
After that, we provide a training session to ensure they understand the task and can use the annotation interfaces (Appendix~\ref{append:claim validation platform} and ~\ref{append: refuted claim validation interface}). We also give them three samples to test their understanding. We recruit twelve annotators that passed the training session. 
% Once they pass the training session, we assign them official batches to work on. 
% \paragraph{Task Design.}
% \paragraph{Payment and Quality Control.}
In compliance with ethical guidelines, we ensure fair compensation for the annotators. 
% We hired a total of 12 annotators for the annotation process. 
Each claim annotation is reimbursed at a rate of 0.37 USD, resulting in an hourly wage of 11.2 USD\footnote{The payment is fair and aligned with the guideline for dataset creation~\cite{DBLP:journals/tacl/BenderF18}.}. 
% It is important to note that the inter-annotator agreement was calculated after the authors' checking process. 

\paragraph{Quality Control and Annotator Agreement.}
% \lxy{combine payment and quality control and inter agreement into 1 paragraph; no need to write kappa definition}
%\lxy{need to clarify the Kappa is calculated by how many data samples, e.g., 900 for NEI and 872 for refuted }
%649 for NEI claim validation; 1225 for refuted claim 
%\lxy{IAA shows that the data is clean, it should be 1/3 of the central argument, but very short now. be more convincing and explaining; helping describe the task definition}
To ensure the quality of the annotation, we apply strict quality control procedures following the guidelines outlined in the Dataset Statement~\cite{DBLP:journals/tacl/BenderF18}. We assign two different annotators to perform a two-round annotation for each claim, while two authors review and resolve any identified errors or issues. To measure the inter-annotator agreement, we use Cohen's Kappa~\cite{Cohen1960ACO}.
% , which takes into account both the observed agreement $p_o$ and the expected agreement by chance $p_e$, and is calculated as $k = \frac{p_o-p_e}{1-p_e}$.
Our inter-annotator agreement is 0.630 for the false claim verification task (872 claims in total) and 0.719 for the NEI claim verification task (900 claims in total). Both values indicate substantial agreement among the annotators. 
% , with 12 annotators for refuted claims and 10 annotators for NEI claims.

\section{Data Analysis}
%\lxy{v2}

Table~\ref{tab:dataset comparison} shows the statistics of our \dataset dataset and the comparison with three existing table fact-checking datasets: TabFact~\cite{DBLP:conf/iclr/ChenWCZWLZW20}, FEVEROUS~\cite{DBLP:conf/nips/AlyGST00CM21}, and SEM-TAB-FACTS~\cite{DBLP:conf/semeval/WangMDR21}. Compared with these datasets, \dataset is 1) annotated by domain experts rather than crowd-sourced workers, 2) contains more challenging claims that require up to 11 reasoning steps for verification, and 3) has a more balanced distribution of veracity labels and a higher percentage of NEI claims. We conduct a more in-depth analysis of \dataset as follows. 

\subsection{Reasoning Analysis}
\label{subsec:reasoning structure}
%\lxy{the most important part}
% To gain a better understanding of the challenges and to show the complexity of the claims presented in \dataset, we conduct an analysis of the types, depths, and program structure of reasoning required to answer the claims. 

% We first conduct an analysis of the reasoning skills required for fact-checking claims in \dataset. 

\paragraph{Reasoning Types.}
To study the nature of reasoning involved in fact-checking claims in \dataset, we adapt the set of table-based reasoning categories from INFOTABS~\cite{DBLP:conf/acl/GuptaMNS20} to define 14 atomic reasoning types, as shown in Table~\ref{tab:symbol}. 
% Appendix~\ref{append:reasoningtab} gives the full list with definitions. 
Among them, ``closed-domain knowledge'' and ``open-domain knowledge'' are specially designed for \dataset. Closed-domain knowledge refers to obtaining background information from the table caption or title, \textit{e.g.}, knowing that ``Prod.'' refers to ``Productivity'' from the table caption in Figure~\ref{fig:data_example}. Open-domain knowledge refers to commonsense knowledge not presented in the table, \textit{e.g.}, the relationship between precision and recall. Given the designed reasoning types, we manually analyze 100 samples in \dataset, by annotating the graph of reasoning steps for verifying each claim. We identify 476 atomic reasoning steps from the 100 analyzed samples and show the proportion for each reasoning type in Table~\ref{tab:symbol}. We observe that \dataset has a multifaceted complex range of reasoning types and a high proportion of claims requiring different types of domain knowledge. 
% (closed-domain, open-domain, commonsense). 

% We adopt and refine the 14 reasoning types (shown in Table~\ref{tab:symbol}) proposed by INFOTABS~\cite{DBLP:conf/acl/GuptaMNS20}, including simple lookup, numerical reasoning, check value, trend same/different, and set check.
% In addition, we have introduced two new types of reasoning, namely %\kmy{Global: please note should be ``Closed'' not ``Close''.} 
% ``Closed-domain knowledge'' and ``Open-domain knowledge'' that are designed specifically for scientific table fact-checking. Closed-domain knowledge refers to information that is available within the paper itself, \textit{e.g.}, from table caption or the article, while Open-domain knowledge refers to domain-specific knowledge that the experts need to know but which is not explained in the article, \textit{e.g.,} BLEU score.
%\kmy{This is tricky, many articles actually explain the formula for $F_1$.}
%By incorporating these two types of reasoning, we aim to improve the accuracy and comprehensiveness of our fact-checking process.
%\lxy{Highlighting the two new types, e.g., 20\% of the datasets, but do not repeat the descriptions. Maybe change to examples in the texts. The key point is to show the different information between the tables and texts}

\begin{table*}[!t]
    \centering
    %\scriptsize
    \resizebox{\textwidth}{!}{
    \begin{tabular}{llc}
    \toprule
         \textbf{Function Names} & \textbf{Descriptions} & \textbf{Prop. (\%)} \\ 
         \midrule
         Simple lookup & Retrieve the value for a specific cell. & 20.6\\ 
         Comparison & Compare two numbers. & 19.5 \\ 
         Closed-domain knowledge & Extract information from context sentences in the table caption or article.  & 12.1\\ 
        Open-domain knowledge & Extract additional information required by domain experts.
%, such as $F_1$, Precision, and Recall. 
& 5.3 \\ 
Commonsense knowledge & Extract commonsense knowledge necessary for claim verification. & 5.3 \\ 
        Subtract & Perform subtraction of two numbers.  & 5.3\\  
         Divide & Perform division of two numbers. & 5.3 \\  
         Rank & Determine the rank of a set of numbers. & 5.3 \\ 
        Different / Same & Determine if two numbers are different or the same.  & 5.3 \\ 
        Add & Calculate the sum of two numbers. & 4.0  \\  
        Max / Min & Retrieve the maximum or minimum number from a set of numbers.  & 3.1\\ 
        Col / Rowname
        %Grounding 
        & Retrieve the column or row name from the table. & 3.1 \\ 
        Trend same/different & Determine the trend for two columns or rows, whether they are the same or different. &2.9 \\
        Set check & Verify if a value belongs to a set of numbers.& 2.9 \\
        \bottomrule
    \end{tabular}
    }
    \caption{The function names, descriptions, and their proportions in our \dataset dataset. 
    %\qian{A little strange to call it in-context knowledge. Maybe close-domain knowledge will be better? For the out-of-context, it can be open-domain knowledge. Greater / Lower may be changed to Comparison. Col / Rowname can be something like Grounding? } 
    }
   % \lxy{change the descriptions to operation}
    \label{tab:symbol}
\end{table*}

\paragraph{Reasoning Depth.}
%Compositional Reasoning
%\lxy{Too long.}
% To present the level of complexity in the claims in our \dataset dataset, we analyze the reasoning depth required to verify the claim. The reasoning depth is measured by the number of individual operations needed in each claim. 
% We visualize the histograms of the reasoning depth distribution in Figure~\ref{fig:reasoning depth}. We observe that the average reasoning step is 4.76 and the maximum reasoning depth is 11. 
% \lxy{shrink this paragraph}
% %Shallow vs Deep claims
% The claims in our \dataset dataset are also classified into two categories based on their level of complexity: shallow claims, which only require one to two reasoning steps, and deep claims, which require three or more reasoning steps. As shown in Figure~\ref{fig:reasoning depth}, the majority of the claims (85\%) fall into the deep claims category, highlighted in blue, while the minority fall into the shallow claims category. % the reasoning steps are concentrated between XXX to XXX. 

We further measure the \textit{reasoning depth} (the number of required reasoning steps) for each claim and show the reasoning depth distribution in Figure~\ref{fig:reasoning depth}. We find that the analyzed claims have an average depth of 4.76 and a maximum depth of 11. Moreover, 86\% of the claims requiring 3 or more reasoning steps, which demonstrates the complexity of reasoning in \dataset. 

% Figure~\ref{fig:reasoning depth} shows the reasoning depth distribution with an average of 4.76 and a maximum depth of 11. The claims are classified into shallow (1--2 reasoning steps) and deep (3 or more steps) categories. Most of the claims (86\%) fall into the deep claims category, as shown in Figure~\ref{fig:reasoning depth}.
%\lxy{expand other information beside the graphs,e.g., compare this dataset to  other dataset}
% This showcases the reasoning difficulty in \dataset.

%%% old figure
% \begin{figure}[!ht]
%     \centering
%     \includegraphics[width=7.5cm]{figures/reasoning_depths.pdf}
%     % \fbox{\rule[-.5cm]{0cm}{6cm} \rule[-.5cm]{6cm}{0cm}}
%     \caption{The distribution histogram of reasoning steps in our \dataset dataset. The x-axis is the reasoning steps in each claim, and the y-axis is the frequency for each reasoning step. The shallow claims (with 1-2 reasoning steps) are highlighted in green, while the deep claims (with 3+ reasoning steps) are highlighted in blue.}
%     \label{fig:reasoning depth}
%   %    \lxy{Larger axis number fonts; percentage and reasoning steps same font}
%     \vspace{-0.3cm}
% \end{figure}

\begin{figure}[tb]
    \centering
    \begin{tikzpicture}[scale=0.95]
    \small{
        \begin{axis}[
                ymajorgrids,
                xmajorgrids,
                grid style=dashed,
                ybar=3pt,
                bar width=.33cm,
                enlarge x limits=0.2,
                width=.5\textwidth,
                height=.3\textwidth,
                symbolic x coords={$1$, $2$, $3$, $4$, $5$, $6$, $7$, $8$, $9$, $10$, $11$},
                xtick={$1$, $2$, $3$, $4$, $5$, $6$, $7$, $8$, $9$, $10$, $11$},
                ytick={5,10,...,25},
                nodes near coords,
                every node near coord/.append style={font=\small},
                nodes near coords align={vertical},
                ymin=0,ymax=25,
                ylabel={Percentages (\%)},
                xlabel={Reasoning Steps}
            ]
             \addplot[red, fill=red!20, bar shift=0pt, text=black] coordinates {
                ($1$, 6)
                ($2$, 8)
            };
             \addplot[blue, fill=blue!20, bar shift=0pt, text=black] coordinates {
                ($3$, 15)
                ($4$, 18)
                ($5$, 20)
                ($6$, 15)
                ($7$, 7)
                ($8$, 5)
                ($9$, 3)
                ($10$, 2)
                ($11$, 1)
             };
        \end{axis}
    }
    \end{tikzpicture}
    \caption{The distribution histogram of reasoning steps in our \dataset dataset. The x-axis is the reasoning steps in each claim, and the y-axis is the frequency for each reasoning step. The shallow claims (with 1--2 reasoning steps) are highlighted in {\textcolor{red}{red}}, while the deep claims (with 3+ reasoning steps) are highlighted in {\textcolor{blue}{blue}}.}
    % \caption{Experimental results in the ablation study. The y-axis is the average metric values across all experimental benchmarks in the privacy protection setting. The backbone model is GBDT.}
    \label{fig:reasoning depth}
\end{figure}

\paragraph{Reasoning Graph.}
%\lxy{Descibing the program structure example in Figure 1}
% To show the reasoning's diversity and complexity of our claims in \dataset, the reasoning graphs are drawn on the right part in Figure \ref{fig:data_example}. In the example, to verify the claim ``A's productivity of 57.5\% expresses that it appears in 7.5\% more often than expected by random chance.'', different types of reasoning and different levels of reasoning paths are involved. For example, first, the `In-context knowledge' from the table caption is used to capture productivity corresponding to Prod. column in the table. Then, `simple lookup' reasoning is contained to find A's productivity is 57.5\%. In another reasoning path, `commonsense knowledge' reasoning is used to interpret random chance as 50\%. Last, two reasoning paths are merged by a `subtraction' reasoning type to calculate the results between 57.5\% and 50\% is 7.5\%. Finally, the claim is supported. 

We showcase the reasoning graph for the example in Figure~\ref{fig:data_example} on the right side of the figure. Verifying this claim requires various types of reasoning including: 1) \textit{background knowledge from the table caption}: ``productivity'' corresponds to the ``Prod.'' column in the table; 2) \textit{commonsense knowledge}: ``random chance'' means 50\% accuracy; 3) \textit{simple lookup}: ``A's productivity'' refers to the cell located at the last row and the ``Prod.'' column; and 4) \textit{numerical reasoning}: the difference between 57.5\% and 50\% is 7.5\%. This case study provides further insights into the complexity and variety of reasoning involved in \dataset, revealing the difficulty of the dataset.

\subsection{Refuted and NEI Claims Analysis}
\label{subsec:refuted and NEI reason}
% For datasets that utilize models for generating claims, it is critical to ensure the diversity of the generated claims. 
One potential risk of model-generated claims is that they may lack diversity and exhibit the same pattern. For example, in the Sci-Fact~\cite{DBLP:conf/emnlp/WaddenLLWZCH20} dataset where the refuted claims are generated by flapping the meaning of the original true claims, we found that out of 100 randomly sampled refuted claims, 85 simply negated the original claim by adding negation words such as ``not'' (more details in Appendix~\ref{append: scifact refuted reason analysis}). To evaluate the diversity of claims for our \dataset dataset, we randomly select 60 refuted claims and then manually annotate their reasons for refutation. Results are shown in Table~\ref{tab:refuted and nei reason} (top half). We find that \dataset exhibits a greater diversity in refuted claims compared to Sci-Fact. Besides common error types such as ``incorrect calculation results'' (41.7\%), there are also unique types of errors that are more reflective of the complexities in real-world scientific claims. For example, 33.33\% of the refuted claims contain ``incorrect approximation words'', and 10.0\% are cases where ``the claim is partially right'', consistent with the fact that ambiguity and half-truths are common phenomena in scientific discourse. Additional examples of refuted claims are in Appendix~\ref{append:real refuted claim case study}. 

% LXY
% One challenge with machine-generated refuted claims is the potential lack of diversity. To assess the diversity of the generated claims in our \dataset,
% we conduct an in-depth analysis by randomly selecting 60 claims from each category. Table~\ref{tab:refuted and nei reason} (top half) presents the distribution of claims in each category based on the type of knowledge required to refute them. Our analysis shows that the distribution of refuted types in our \dataset dataset is consistent with other fact-checking datasets, such as incorrect calculation results (41.7\%), values that do not match (8.3\%), and wrong operation type (6.7\%). More examples of refuted claims are shown in Appendix~\ref{append:real refuted claim case study}.

% However, two unique refuted types in our \dataset dataset are incorrect approximation words (33.33\%) and the claim is partially right (10.0\%). These findings indicate that our dataset is more representative of real-world scientific claims, where ambiguity and half-truth cases are frequent in scientific fact-checking. In order to provide a relevant comparison, we selected the Sci-Fact~\cite{DBLP:conf/emnlp/WaddenLLWZCH20} dataset, which also focuses on fact-checking in the scientific domain. We analyzed the refuted reasons in their dataset, as shown in Appendix~\ref{append: scifact refuted reason analysis}. In contrast to Sci-Fact, where simple negation (+not) accounts for over 85\% of the refuted reasoning, our dataset demonstrates a greater diversity in the reasons for refutation.

The NEI claims (bottom half; Table~\ref{tab:refuted and nei reason}) also exhibit diverse reasoning patterns. The two most common features for unverifiable claims are insufficient evidence in the table and the lack of background knowledge. The lack of closed-domain knowledge is another reason for NEI, where additional information in the paper is necessary to verify the claim. Other reasons include the use of vague pronouns (\textit{e.g.}, ``it'', ``this'') brings ambiguity to the claim. These distinct refuted and NEI reasoning types highlight the unique features of \dataset, making it a more comprehensive and realistic representation of the challenges faced in real-world scientific fact-checking. 

% For example, except for not having enough matching evidence, open-domain knowledge is necessary for claims that require familiarity with specific terminologies such as BLEU. Lack of closed-domain knowledge is another reason for NEI, where a clear understanding of the model architecture mentioned elsewhere in the paper is necessary to verify the claim. Other reasons include the use of vague pronouns (\textit{e.g.}, ``it'', ``this'') brings ambiguity to the claim. These distinct refuted and NEI reasoning types underscore the unique features of \dataset, making it a more comprehensive and realistic representation of the challenges faced in real-world scientific fact-checking.
%%%%

\begin{table}[!t]
\centering
%\scriptsize
\resizebox{0.48\textwidth}{!}{
\begin{tabular}{lc}
\toprule
\textbf{Refuted Reasons}& \textbf{Prop. (\%)}\\\toprule
The calculation result is wrong. & 41.7\\
The approximation word is wrong. & 33.3\\
The claim is partially right. & 10.0\\
The values in the claim do not match. & 8.3\\
The operation type is wrong. & 6.7\\
\midrule
% \midrule
\textbf{NEI Reasons}& \textbf{Prop. (\%)}\\\midrule
The claim does not have enough matching evidence. & 33.3\\
The claim lacks open-domain knowledge. & 25.0\\
The claim lacks closed-domain knowledge. & 15.0\\
The claim refers to another table. & 11.7\\
The claim contains vague pronouns. & 8.3\\
The claim omits specific information. & 6.7\\
\bottomrule
\end{tabular}
}
\caption{Refuted and NEI reasons and their estimated proportions (Prop.) in \dataset. 
%\qian{The font size is relatively small.}
}
\label{tab:refuted and nei reason}
\vspace{-0.3cm}
\end{table}

%%%Old Table
% \begin{table}[!t]
% \centering
% %\scriptsize
% \resizebox{0.49\textwidth}{!}{
% \begin{tabular}{clc}
% \toprule
% & \textbf{Reasons}& \textbf{Prop. (\%)}\\ \toprule
%  & The calculation result is wrong. & 41.7\\
% \textbf{Refuted} & The approximation word is wrong. & 33.3\\
% \textbf{Claims}&The claim is partially right. & 10.0\\
% &The values in the claim do not match. & 8.3\\
% &The operation type is wrong. & 6.7\\
% \midrule
% & The claim does not have enough matching evidence. & 33.3\\
% & The claim lacks out-of-context knowledge. & 25.0\\
% \textbf{NEI}& The claim lacks in-context knowledge. & 15.0\\
% \textbf{Claims} & The claim refers to another table. & 11.7\\
% & The claim contains vague pronouns. & 8.3\\
% & The claim omits specific information. & 6.7\\
% \bottomrule
% \end{tabular}
% }
% \caption{Refuted and NEI reasons and their proportions in \dataset. \qian{The font size is relatively small.}}
% \label{tab:refuted and nei reason}
% \vspace{-0.3cm}
% \end{table}

%Modeling
\section{Experiment}
% In this study, we examine the extent to which pre-trained language models (LMs) learn to classify the fact verification task that involves tables. We evaluate several methods in our \dataset dataset in Section~\ref{sec:baselines}. We show the main results in Section~\ref{sec:main results} and error analysis in Section~\ref{sec:error analysis}.

% We then conduct a comprehensive evaluation of the state-of-the-art models on \dataset. 

%%%%Main Results
\begin{table*}[!t]
\centering
\resizebox{1\textwidth}{!}{
\renewcommand{\arraystretch}{1.0}
\begin{tabular}{clrcccc}
\toprule
  \multicolumn{2}{c}{\multirow{2}{*}{\textbf{Models}} } & 
  \multirow{2}{*}{\textbf{\# of Para.}} &
  \multicolumn{2}{c}{\textbf{Zero-shot}} & 
  \multicolumn{2}{c}{\textbf{In-Context}} 
  \\  
  \cmidrule(lr){4-5} \cmidrule(lr){6-7} 
  %\cmidrule(lr){8-9} \cmidrule(lr){10-11} \cmidrule(lr){12-13}
 & & & \textbf{2-class} & \textbf{3-class} & \textbf{2-class} & \textbf{3-class}  
 %& \textbf{2-class} & \textbf{3-class} & \textbf{2-class} & \textbf{3-class} & \textbf{2-class} & \textbf{3-class}
 \\ \midrule
\multirow{4}{*}{} 
 & \texttt{TAPAS-large (Tabfact)} \small{~\cite{DBLP:conf/acl/HerzigNMPE20}} & 340M&50.30 & --- & --- & ---\\
\uppercase\expandafter{\romannumeral1.} Table-based & \texttt{TAPEX-large (Tabfact)} \small{~\cite{DBLP:conf/iclr/LiuCGZLCL22}}&400M& 56.06 & --- & --- & ---\\
 LLMs  & \texttt{TAPEX-Zero-large} \small{~\cite{liu2023zero}} & 780M & 48.28 & 29.72 & 42.44 & 23.47\\
& \texttt{TAPEX-Zero-XL} \small{~\cite{liu2023zero}} & 3B & 49.77 & 34.30 & 42.12 & 25.62\\
\midrule
\multirow{5}{*}{} 
& \texttt{Flan-T5-base} \small{~\cite{DBLP:journals/corr/abs-2210-11416}}&250M & 47.38  & 26.56 & 44.82 & 24.09 \\
\uppercase\expandafter{\romannumeral2.} Encoder--Decoder & \texttt{Flan-T5-large} \small{~\cite{DBLP:journals/corr/abs-2210-11416}}&780M & 51.58 & 32.55 & 49.62 & 27.30\\
LLMs & \texttt{FLan-T5-XL} \small{~\cite{DBLP:journals/corr/abs-2210-11416}} &3B & 52.41
& \textbf{38.05} & 48.05 & 29.21 \\
& \texttt{Flan-T5-XXL} \small{~\cite{DBLP:journals/corr/abs-2210-11416}}&11B & 59.60 & 34.91 & \textbf{60.48} & 34.04 \\
% & \texttt{Flan-Alpaca-X}\small{~\cite{flan-alpaca}}&X & X & X & X & X \\
 \midrule
\multirow{6}{*}{} 
& \texttt{Alpaca-7B} \small{~\cite{alpaca}} &7B & 37.22 & 27.59 & 40.46 & 28.95 \\
\uppercase\expandafter{\romannumeral3.} Open source & \texttt{Vicuna-7B}\small{~\cite{vicuna2023}}& 7B & \textbf{63.62} & 32.47 & 50.35 & 34.26 \\
LLMs& \texttt{Vicuna-13B}\small{~\cite{vicuna2023}} & 13B & 41.82 & 29.63 & 55.11 & \textbf{35.16} \\
& \texttt{LLaMA-7B} \small{~\cite{touvron2023llama}} & 7B & 49.05 & 32.26 & 45.24 & 27.17 \\
& \texttt{LLaMA-13B} \small{~\cite{touvron2023llama}} & 13B & 53.97 & 37.18 & 44.39 & 32.66  \\
\midrule
\multirow{5}{*}{}
  & \texttt{InstructGPT}\small{~\cite{ouyang2022training}} & 175B & 68.44 & 41.41 & 68.10 & 41.58 \\
 \uppercase\expandafter{\romannumeral4.} Close source   &\texttt{InstructGPT+CoT}\small{~\cite{ouyang2022training}} & 175B & ---& --- & 68.46 & 42.60 \\
LLMs& \texttt{PoT} \small{~\cite{chen2022program}}  &175B & --- & --- & 63.79 & --- \\
& \texttt{GPT-4} \small{~\cite{DBLP:journals/corr/abs-2303-08774}} & --- & \underline{78.22} & \underline{64.80 }& \underline{77.98} & \underline{63.21} \\
& \texttt{GPT-4+CoT} \small{~\cite{DBLP:journals/corr/abs-2303-08774}} & --- & --- & --- & 76.85 & 62.77 \\
\midrule 
 & Human  &--- &  --- & --- & 92.40& 84.73 \\
 \bottomrule
\end{tabular}%
}
% \caption{Macro-$F_1$ scores of baselines on \dataset for different settings. The TAPAS-large and TAPEX-large models are fine-tuned on the TabFact dataset, while others perform zero-shot learning.  The bold text indicates the best performance from \uppercase\expandafter{\romannumeral1} to \uppercase\expandafter{\romannumeral4}, and underlined text indicates the best from \uppercase\expandafter{\romannumeral1} to \uppercase\expandafter{\romannumeral3}.}  
\caption{Macro-$F_1$ of baselines on \dataset for different settings. The \textit{\# of Para.} indicates the number of parameters in the models. The TAPAS and TAPEX models are fine-tuned on the TabFact dataset, while others perform zero-shot learning. 
% \uppercase\expandafter{\romannumeral4} represents API-based LLMs, which are black-box, while \uppercase\expandafter{\romannumeral3} corresponds to decoder-only LLMs, which are open source and replicable. 
The bold text indicates the best performance among \uppercase\expandafter{\romannumeral1} to \uppercase\expandafter{\romannumeral3}, while the underlined text indicates the overall best performance among all the models. }
%\lxy{Add Fine-tuned open-source LLMs}
\label{tab:main result}
\end{table*}
%\underline{\color{darkgreen} \textbf{67.66}}
% \subsection{Experimental Settings}

% \subsection{Task Formulation}

We formally define the task of scientific table-based fact-checking as follows. A scientific table $\mathcal{T}$ consists of a table caption $P$ and the table content $(\{T_{i,j} \vert i \leq R_T, j \leq C_T \}$ with $R_T$ rows and $C_T$ columns, where $T_{i,j}$ is the content in the $(i, j)$th cell. Given a claim $C$ describing a fact to be verified against the table $\mathcal{T}$, a \textit{table fact-checking} model $\mathcal{F}$ predicts a label $\mathcal{Y}$ to verify whether $\mathcal{C}$ is supported, refuted, or can not be verified by the information in $\mathcal{T}$. 

Considering the real-world situation that large-scale training data is either not available or expensive to collect, we focus on the \textit{zero-shot/in-context} evaluation where the model can only access zero/few in-domain data from \dataset. To this end, we randomly hold out 5 tables with 25 claims as model-accessible data and use the rest of the data as the unseen test set. This also prevents the model from learning spurious features that lead to over-estimated performance~\cite{DBLP:conf/emnlp/SchusterSYFSB19}. 

\subsection{Models}
\label{sec:baselines}
We conduct a comprehensive evaluation of \dataset for various models, including table-based pretraining models, encoder–decoder models, open source LLMs, and closed source LLMs. We also study the human performance to analyze the upper bounds on \dataset. 
%\lxy{[DONE]show that API-based LLMs are not replicable, it is a closed-model, and the Decoder-only LLMs are open source, which is replicable.} 
% Notably, API-based LLMs are black-box models, making them non-replicable, whereas decoder-only LLMs are open source and offer replicability for further research. 
\vspace{-0.1cm}

\paragraph{Table-based LLMs.} These are pre-trained transformer models fine-tuned on tabular data. We choose three different models: 1) \texttt{TAPAS}~\cite{DBLP:conf/acl/HerzigNMPE20}, a BERT-based model fine-tuned on millions of tables from English Wikipedia and corresponding texts, 2) \texttt{TAPEX}~\cite{DBLP:conf/iclr/LiuCGZLCL22}, a model that fine-tunes BART~\cite{lewis2019bart} on a large-scale synthetic dataset generated by synthesizing executable SQL queries and their execution outputs, and 3) \texttt{TAPEX-Zero}~\cite{liu2023zero}, an enlarged version of \texttt{TAPEX}. For \texttt{TAPAS} and \texttt{TAPEX}, we use their fine-tuned version on TabFact~\cite{DBLP:conf/iclr/ChenWCZWLZW20} for table fact-checking. 
\vspace{-0.1cm}

\paragraph{Encoder–Decoder LLMs.} We also use encoder–decoder models where both the input and output are sequences of tokens. To adapt the model to take the table as input, we flatten the table as a sequence following~\citet{DBLP:conf/iclr/ChenWCZWLZW20}. The input is then formulated as $[\tilde T; P; C; Q]$, where $\tilde T $ is the linearized table, and $Q$ is a question template ``Based on the information in the table, is the above claim true? A) True B) False C) Unknown?''. We choose \texttt{FLAN-T5}~\cite{DBLP:journals/corr/abs-2210-11416}, an improved T5 model~\cite{DBLP:journals/jmlr/RaffelSRLNMZLL20} pre-trained on more than 1.8K tasks with instruction tuning, which has achieved strong zero-shot/in-context performance on other fact-checking benchmarks. 
%We also use \texttt{Flan-Alpaca}~\cite{flan-alpaca}, a model extending the instruction tuning of Stanford Alpaca~\cite{alpaca} to FLAN-T5. 
\vspace{-0.1cm}

\paragraph{Open Source LLMs.} 
% \lxy{shrink because}
We also evaluate the performance of state-of-the-art open source LLMs, including 1) \texttt{LLaMA}~\cite{touvron2023llama}, the first open-source model by Meta AI; 2) \texttt{Alpaca}~\cite{alpaca}, an instruction-following language model fine-tuned on LLaMA; and 3) \texttt{Vicuna}~\cite{vicuna2023}, the arguably best-performed open-source LLMs that claimed to achieve 90\% quality compared to OpenAI \texttt{ChatGPT}. 
% and Google \texttt{Bard}. 
We use the same input format as in the encoder-decoder model. 
\vspace{-0.1cm}

\paragraph{Closed Source LLMs.} These are closed-source LLMs that require API calls for inference, including \texttt{InstructGPT (text-davinci-003)}~\cite{ouyang2022training} and \texttt{GPT-4}~\cite{DBLP:journals/corr/abs-2303-08774}. We evaluate the setting that directly predicts the label and the \texttt{Chain-of-Thought} (CoT)~\cite{DBLP:CoT} setting, which generates explanations before predicting the final label. We also include the \texttt{Program-of-Thoughts} (PoT)~\cite{chen2022program} model that has shown strong ability in solving complex numerical reasoning tasks. It first parses the reasoning steps as Python programs and then executes them on a Python interpreter to derive accurate answers. Since most claims in \dataset also require numerical reasoning, we want to test whether program-guided reasoning can be extended to table-based fact-checking. 
\paragraph{Human Performance.} To examine how humans perform on our \dataset dataset, we hired an annotator from our candidate annotators pool, following the same training procedure as other annotators. In the case of 2-class classification, we randomly selected 40 samples: 20 each for supported and refuted claims. For 3-class classification, we randomly selected 60 random samples, ensuring an even distribution of 20 samples across the three label categories (supported, refuted, and not enough information). The annotator took approximately 1.5 hours for the 2-class fact-checking task and 2 hours for the 3-class setting. We report the Macro-F1 scores at the bottom of Table~\ref{tab:main result}. 

\subsection{Main Results}
\label{sec:main results}
%\lxy{Global: Few-shot-> In-Context setting}
We evaluate all models under both \textit{zero-shot} and \textit{in-context} settings. In the zero-shot setting, the model does not have access to any in-domain data. In the in-context setting, we provide three hold-out examples as demonstrations. We report two sets of results: the \textit{2-class} case, where examples labeled as NEI are excluded (since some models cannot process NEI claims), and the \textit{3-class} case including all three labels. The results are shown in Table~\ref{tab:main result}. We have five major observations. 
% \lxy{order the observations from the most important one to the least important one, currently 2 is showing that GPT-4 is kind of solving the dataset [order with 1,3,4,5,2]}

\ul{1. In general, all open source LLMs, including encoder--decoder models and decoder-only models, do not achieve very promising results on SCITAB and they still have a large gap from human performance.} The best result is 63.62 for the 2-class setting (\texttt{Vicuna-7B} and 38.05 for the 3-class setting (\texttt{FLAN-T5-XL}). Both results are only moderately better (+13.62 and +4.72) than random guessing. In contrast, a well-trained human annotator can achieve 92.46 and 84.73 F1 scores in the 2-class and 3-class settings, respectively. This reveals the challenging nature of \dataset and its potential to be the future benchmark for scientific fact-checking. 

\ul{2. Counter-intuitively, table-based LLMs do not outperform models pre-trained on pure texts}, for example, \texttt{FLAN-T5}. This discrepancy may be attributed to the dissimilarity between the distribution of tables in scientific literature and publicly available table corpus. For example, scientific tables commonly include both row and column headers, whereas most tables in Wikipedia lack row headers. Meanwhile, the claims in our dataset are usually much longer than those in previous works, raising challenges to table-based LLMs.

\ul{3. The results in the 3-class setting are notably poorer than those in the 2-class setting.} This discrepancy reveals the challenges that most models face when confronted with the NEI class. One plausible explanation could be the inherent difficulty in distinguishing between `refuted' and `NEI' claims --- a task that even trained human annotators struggle with, as noted by~\citet{DBLP:conf/emnlp/JiangBZD0B20}. Our forthcoming error analysis will further demonstrate that the inclusion of the NEI class tends to diminish the models' confidence, causing a shift in their predictions from `supported/refuted' to `NEI'. 

\ul{4. Interestingly, the provision of in-context examples does not result in improved performance for the majority of models.} This observation is somewhat expected for open source LLMs as they have not been reported to possess in-context learning capabilities. Nonetheless, it is surprising to find that even with chain-of-thought prompting, in-context demonstrations do not yield positive effects for InstructGPT and GPT-4. Our error analysis on the PoT offers some insight into this phenomenon and will be discussed in the next section. 

\ul{5. Closed source LLMs perform better than open source LLMs}, with GPT-4 achieving 78.22 macro-$F_1$ for the 2-class setting and 64.80 for the 3-class setting. This aligns with the assertion that GPT-4 has a strong ability to perform complex reasoning~\cite{DBLP:journals/corr/abs-2303-08774} and we show that this ability can generalize to tabular data as well. However, the black-box nature of OpenAI models restricts our further analysis of its behavior. 

\subsection{Error Analysis}
\label{sec:error analysis}

% To better understand the challenges of our \dataset dataset, we conduct an error analysis for InstructGPT, GPT-4, and PoT. 

\begin{figure}[!t]
    \centering
    \includegraphics[width=7.6cm]{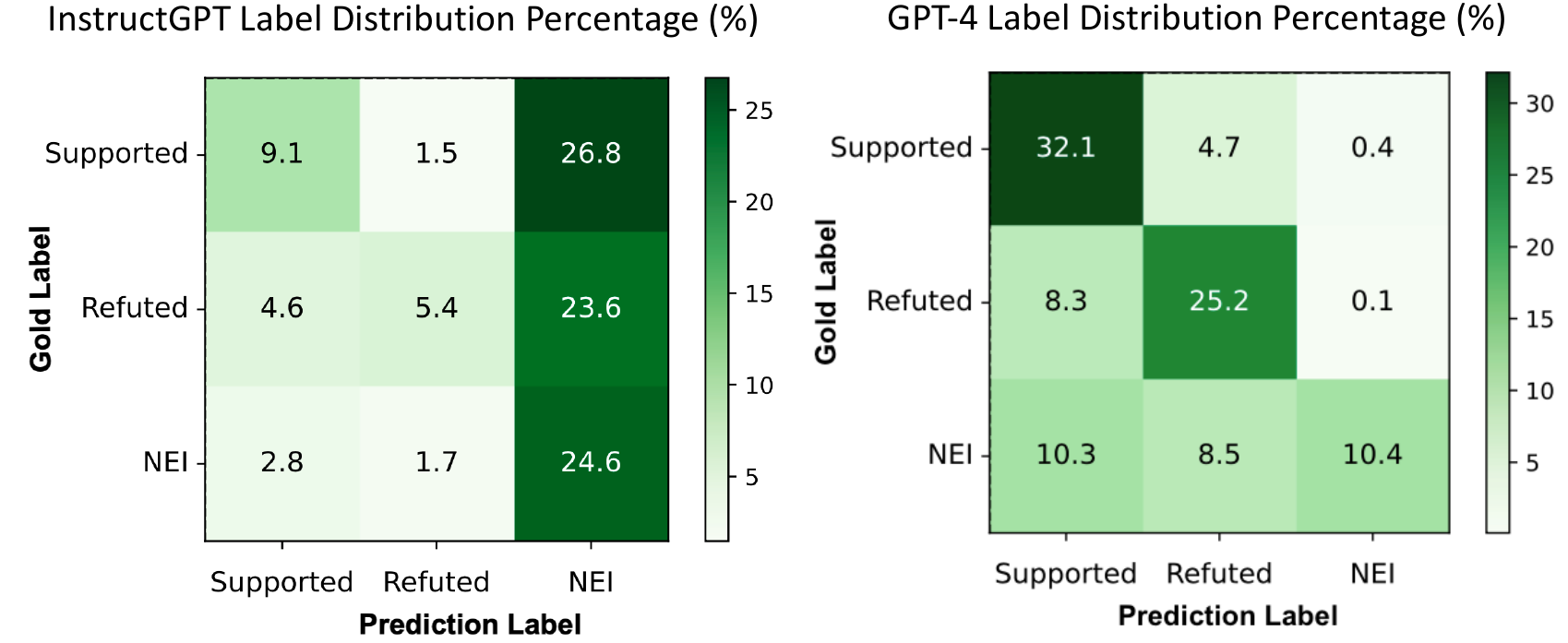}
    \caption{Confusion matrix for InstructGPT (left) and GPT-4 (right) in the zero-shot 3-class classification task.}
    \label{fig:GPT3.5cm}
\end{figure}

\paragraph{InstructGPT and GPT-4.} We show the confusion matrices for InstructGPT and GPT-4 under the zero-shot 3-class setting in Figure~\ref{fig:GPT3.5cm}. We find that both models have difficulty in accurately predicting the NEI class. InstructGPT displays a pattern of ``less confident'', frequently classifying supported and refuted claims as `NEI'. In contrast, GPT-4 exhibits overconfidence, incorrectly categorizing NEI claims as either supported or refuted. This corroborates our earlier observation that distinguishing whether a claim is \textit{verifiable} is one of the key challenges for \dataset. 

Further, we also examine individual error instances, with typical examples provided in Figures~\ref{fig:GPT3.5error_1} and~\ref{fig:GPT3.5error_2} of Appendix~\ref{append:error cases for GPT3.5}. The majority of `supported' claims that were incorrectly classified as `refuted' (Case~6) involve numerical reasoning or comparison. Conversely, when `refuted' claims are inaccurately predicted as `supported' (Case~3), we find that LLMs often overlook claims containing negation, 
%\lxy{put citation of lLMs does not pay attention of negation.}
indicating a lack of deep comprehension. For cases where `supported' or `refuted' claims are erroneously predicted as `NEI' (Cases~1 and 2), such claims typically demand extensive reasoning and a deep understanding of the research findings. Interestingly, when faced with these complex cases, the model tends to default to the safer choice of `uncertain' (NEI). 
% This underscores the importance of NEI claims and highlights the more balanced distribution of NEI claims in our \dataset dataset.
% \lxy{ connecting back to the proportion of NEI, highlighting the need for NEI.}

\paragraph{PoT.}
Unexpectedly, incorporating a Python interpreter does not confer any advantage on our dataset (as shown in Table~\ref{tab:main result}), despite its positive impacts on other numerical reasoning tasks. In order to understand this, we randomly selected 50 claims wherein the PoT incorrectly predicted the final veracity labels and evaluated the quality of the generated Python programs. We divide the errors into four categories, as assessed by human annotators: (\emph{i})~\textit{Grounding errors}, where the program incorrectly associates data with the respective cells in the table; (\emph{ii})~\textit{Ambiguity errors}, where the claim contains ambiguous expressions that the program fails to represent;  (\emph{iii})~\textit{Calculation errors}, where incorrect floating point arithmetic calculation in Python lead to inaccurate results and (\emph{iv}) \textit{Program errors}, which encompass mistakes such as incorrect or missing arguments/variables, and erroneous operations. We present the error analysis in Table~\ref{tab:pot_error_analysis}, and examples of program errors can be found in Figure~\ref{fig:PoTerror_1} and Figure~\ref{fig:PoTerror_2} in Appendix~\ref{append:pot_error}. 
Compared to other datasets, categories (\emph{i}) and (\emph{ii}) present unique challenges in our dataset. Category~(\emph{i}) underlines the difficulty in accurately referencing the specific cells to which a claim refers. Category~(\emph{ii}), on the other hand, emphasizes the difficulties posed by the ambiguous nature of scientific claims, such as ``A is significantly better than B'', to program-based methods. This connection further emphasizes the contribution of our work in addressing the mismatches between reasoning types and the occurrence of grounding errors.

\begin{table}[!t]
\centering
\resizebox{\columnwidth}{!}{
\begin{tabular}{lc}
\hline
\textbf{Error Type}& \textbf{Estimated Proportion (\%)}\\ \hline
{\uppercase\expandafter{\romannumeral1}}. Grounding errors & 50\\
% \quad \quad a. Wrong entity linking & \quad \quad 18\\
% \quad \quad b. Incomplete entity linking & \quad \quad 32\\ 
%\hline
{\uppercase\expandafter{\romannumeral2}}. Ambiguity errors & 22\\
% \quad \quad a. Wrong operation  & \quad \quad 8 \\
% \quad \quad b. Wrong variables & \quad \quad 0\\ 
%\hline
{\uppercase\expandafter{\romannumeral3}}. Calculation errors & 20 \\ 
%\hline
 {\uppercase\expandafter{\romannumeral4}}. Program errors &  8 \\
\hline
\end{tabular}
}
    \caption{The error types and their estimated proportions for incorrectly-predicted samples in PoT.}
    \label{tab:pot_error_analysis}  
\end{table}

%Related Works
\section{Related Work}
%1. What is FV? 2. Most FV datasets are based on text, such as XXX, 3. There are also some table-based FV 4. but our dataset is different from theirs. (summarize 3.1, but different from it)

% \kmy{overly repetitive.  This paragraph part may not be very needed due to the introduction already highlighting these areas.  I suggest that you work in the rest of this section into the intro and say at the beginning, that aside from fact checking (already covered) you need to cover related work in SDP. (actually this may not be needed as well)}
% fact-checking datasets.

\paragraph{Scientific Fact-Checking Datasets.}
Existing datasets for scientific fact-checking are summarized in a recent survey from~\citet{vladika2023scientific}. These datasets differ in: 1) \textit{domain}: biology~\cite{DBLP:conf/emnlp/WaddenLLWZCH20,DBLP:conf/naacl/AkhtarCS22}, COVID-19~\cite{DBLP:conf/acl/SaakyanCM20,sarrouti-etal-2021-evidence-based,mohr-etal-2022-covert,wang2023checkcovid}, and climate~\cite{DBLP:journals/corr/abs-2012-00614}, 2) \textit{claim creation}: crowd-sourced claims \textit{v.s.} natural claims, and 3) \textit{evidence source}: Wikipedia articles~\cite{DBLP:journals/corr/abs-2012-00614} or research papers~\cite{DBLP:conf/emnlp/WaddenLLWZCH20, DBLP:conf/emnlp/WaddenLKCBWH22, sarrouti-etal-2021-evidence-based}. However, most of these datasets rely on text evidence to verify claims. SEM-TAB-FACTS~\cite{DBLP:conf/semeval/WangMDR21} is the only existing dataset based on scientific tables, but it is limited to simple, crowd-sourced claims. To bridge this gap, we construct \dataset which contains complex claims from authentic scientific papers with table-based evidence. 

% Various \textit{fact-checking datasets} have been proposed, including human-crafted claims based on Wikipedia~\cite{DBLP:conf/naacl/ThorneVCM18,DBLP:conf/lrec/SatheALPP20,DBLP:conf/naacl/SchusterFB21} and naturally occurring claims in the political and scientific domain~\cite{DBLP:conf/acl/SaakyanCM20,DBLP:conf/acl/GuptaS20,DBLP:conf/emnlp/WaddenLLWZCH20, DBLP:conf/emnlp/GuF0NZ022,DBLP:conf/emnlp/WaddenLKCBWH22,wadden-etal-2022-multivers,wright-etal-2022-generating}. Most of these datasets rely on text evidence to verify claims. Among those few table-based fact-checking datasets, TabFact~\cite{DBLP:conf/iclr/ChenWCZWLZW20} contains Wikipedia tables as evidence for natural language statements, while FEVEROUS~\cite{DBLP:conf/nips/AlyGST00CM21} uses Wiki articles and annotates evidence in the form of sentences or cells from tables. SEM-TAB-FACTS~\cite{DBLP:conf/semeval/WangMDR21} is the only existing dataset for scientific table fact verification, but it is limited to simple, crowd-sourced claims. To address these limitations, we construct a new dataset, \dataset, that contains complex claims from original scientific papers with table-based evidence. 

\paragraph{Table-based Reasoning.} Table-based reasoning requires reasoning over both free-form natural language queries and (semi-)structured tables. Early works either rely on executable languages (\textit{e.g.}, SQL and SPARQL) to access the tabular data~\cite{yin2016neural,yu-etal-2018-spider} or employ graph neural networks to capture logical structure in statements, \textit{e.g.}, LogicFactChecker~\cite{zhong2020logicalfactchecker} and ProgVGAT~\cite{yang-etal-2020-program}. However, these approaches often struggle with generalization, as they are tightly bound to specific table formats and language patterns. To address this, we have seen a shift toward table pre-training, with the advent of Table-BERT~\cite{DBLP:conf/iclr/ChenWCZWLZW20}, TAPAS~\cite{DBLP:conf/acl/HerzigNMPE20}, SaMoE~\cite{zhou2022tablebased}, PASTA~\cite{DBLP:conf/emnlp/GuF0NZ022}, and DATER ~\cite{ye2023large}. These methods encode sentence-table pairs using language models and transform table-based reasoning into question-answering or natural language inference. In our work, we focus on evaluating pre-training-based methods on \dataset because they not only demonstrate superior performance but also offer the benefits of few-shot learning. 

\section{Conclusion and Future Work}
We present \dataset, a novel dataset for scientific fact-checking that addresses the limitations of existing benchmarks. By incorporating real-world scientific claims and their corresponding evidence in the form of tables, \dataset offers a more comprehensive and fine-grained representation of scientific reasoning. The challenging nature of \dataset is evident from the performance of the state-of-the-art, highlighting the need for further research. For example, we believe that addressing the challenges posed by ambiguous claims represents a crucial direction for research in scientific fact-checking~\cite{glockner2023ambifc,liu2023were}. One potential approach is to enhance the disambiguation of ambiguous claims by leveraging contextual information or external knowledge sources. Additionally, studying the compositionality in table-based reasoning is an interesting direction. Consider the work of Self-Ask~\cite{DBLP:journals/corr/abs-2210-03350}, which proposed the ``compositionality gap'' metric to measure the capability of LLMs in compositional reasoning. Such evaluations can be enriched by annotating \dataset with ground-truth reasoning depths and structured reasoning graphs. Beyond this, another direction worth exploring is equipping the LLMs with external tools to further improve the model. For example, the use of GPT-4 plugins, Program-guided Fact-Checking~\cite{DBLP:conf/acl/PanWLLWKN23} or adopting approaches from other tool-augmented LLMs like Toolformer~\cite{DBLP:journals/corr/abs-2302-04761} and Chameleon~\cite{DBLP:journals/corr/abs-2304-09842}.

% One potential approach is to explore methods that enhance the understanding and disambiguation of ambiguous claims. This could involve leveraging contextual information from the surrounding text, utilizing external knowledge sources, or incorporating domain-specific ontologies to provide additional context and to disambiguate the claim. 
%Last, in the case of NEI claims, developing techniques to improve the identification and handling of such claims is essential. This could involve exploring methods that capture and model uncertainty, as well as techniques that utilize multiple pieces of evidence or reasoning paths to make more informed judgments.

\section*{Ethics Statement}
We have received approval from the Institutional Review Board (IRB)\footnote{\url{https://www.nus.edu.sg/research/irb}. The NUS-IRB Reference Code is NUS-IRB-2022-599} for our data collection. The IRB reviewed our experimental design and research procedures to ensure that they do not pose more than minimal risks to research participants. We take steps to protect research participants' privacy and the confidentiality of their data. The review process took two months to complete.

\section*{Limitations}
Firstly, the method and dataset are primarily designed for languages with limited morphology, such as English. Secondly, our \dataset dataset is specifically focused on fact-checking scientific claims based on tables, which represents only one aspect of scientific fact-checking. Further research can explore the integration of other forms of evidence, including textual evidence and figure evidence, to enhance the fact-checking process. Thirdly, our \dataset dataset is primarily focused on numerical reasoning types, as it is derived from the SciGen dataset, which also emphasizes numerical reasoning. It would be beneficial for future studies to incorporate a wider range of reasoning types to provide a more comprehensive fact-checking framework. Lastly, it would be valuable to explore additional annotation types, such as reasoning graphs, to further enrich the depth of analysis and capture more intricate relationships within the claims and evidence.

\section*{Acknowledgements}
This research is supported by the Ministry of Education, Singapore, under its MOE AcRF Tier 3 Grant (MOE-MOET32022-0001). The computational work for this article was partially performed on resources of the National Supercomputing Centre, Singapore (\url{https://www.nscc.sg}).
\bibliography{custom}
\bibliographystyle{acl_natbib}

\clearpage

\appendix
\label{sec:appendix}
%%%STEP1

\section{Claim Extraction Procedure}
\subsection{Claim Definition}
\label{append: claim definition}
In academic writing~\cite{lee2009research},
the accompanying text for data, presented as tables and figures), typically includes three fundamental elements as outlined below. These elements encompass the definition of claims, which involve highlighting key data (KD) and commenting on key data (COM) that emphasizes and comments on the key data.

\paragraph{Location of results (LOC).}Statements that locate where the figure/table is found, \textit{e.g.,}  Figure 7 displays the mean percentile scores. 
\paragraph{Highlighting of key data (KD).}Statements that highlight the important data, \textit{e.g.,}(1) Highest or lowest values (2) Overall trend or pattern in the data (3) Points that do not seem to fit the pattern or trend, etc. (4) Results which provide answers to your research questions 
\paragraph{Commenting on key data (COM).}Statements that interpret the data. There are three types of comments:  (1) Generalization (deductions and implications drawn from the results), \textit{e.g.,} ``This indicates that ...'' (2) Comparison of results with those from prior studies, \textit{e.g., }``Different from ...''
(3) Explanation or speculation (possible reasons or cause-effect relationships for the results), \textit{e.g.,} ``The possible reason is that ...''

\subsection{Claim Extraction Interface}
\label{append:claim classification interface}
\begin{figure*}[!tbh]
\centering
    \includegraphics[width=15cm]{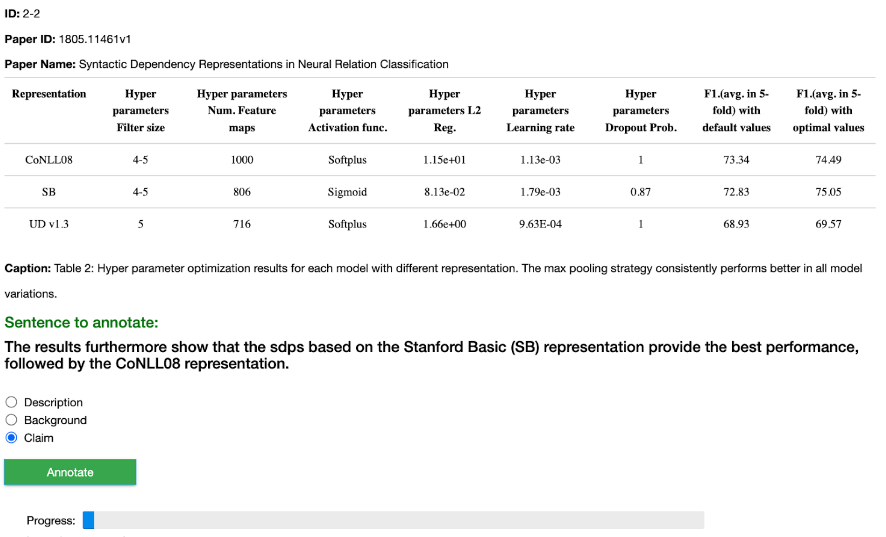}
    \caption{The user interface for the claim extraction task.}
    \label{fig:claim classification ui}
\end{figure*}

Figure~\ref{fig:claim classification ui} shows the user interface for the claim extraction task.

% \newpage

%\subsection{Hiring Questionnaire}

% \subsection{Automatic Classification Results}
% \label{append:bert results}
% By using BERT~\cite{DBLP:conf/naacl/DevlinCLT19} classifier, the 2-class classification and 3-class classification results are shown in Table~\ref{tab:bert result}.

% \begin{table}[!ht]
% \centering
% \resizebox{0.5\textwidth}{!}{
% \begin{tabular}{ccc}
% \hline
% \textbf{Tasks}& \textbf{Predict Accuracy} & \textbf{Macro $F_1$} \\ \hline
% 2-class & 0.8814   & 0.8535 \\
% 3-class & 0.8376  & 0.7315 \\
% \hline
% \end{tabular}
% }
% \caption{BERT in claim classification tasks. 2-class task contains the labels of \{\textbf{Claim, Non-claim}\}, while the 3-class tasks contain the labels of \{\textbf{Descriptions, Background, Claims}\} 
% }
% \label{tab:bert result}
% \vspace{-0.3cm}
% \end{table}

%%%STEP2
\section{Manual Claim Verification Procedure}
% The human verifies the generated NEI claims and the generated Refuted claims.
\subsection{Annotator Training Process}
Our annotator selection and training process is systematic and thorough to ensure the highest quality annotations. We initiate the process by advertising on our university's platform. Interested candidates are then required to complete a registration form. From these responses, the authors identify suitable annotators based on set criteria. Once shortlisted, the potential annotators are invited for a training session, which can be conducted either in-person or via Zoom, lasting approximately one hour. This session is divided into three parts. Firstly, the authors provide a comprehensive overview of the task definition, ensuring clarity on what is expected. Similar to WANLI~\cite{DBLP:conf/emnlp/LiuSSC22}, during our training sessions\footnote{All the related materials including the advertisement, a sample of the registration form and the agreement sheet are available at \url{https://github.com/XinyuanLu00/SciTab}.}, commonsense interpretations and a minimum amount of logical inference are acceptable. Next, a demonstration is given on how to navigate and utilize the annotation interface effectively. Following this, a series of trial tests are released to the annotators. This is to verify their understanding and capability in the task. Last, we specify the deadline for completing annotations, outline how we check the quality of their work, brief them on a post-annotation survey, and explain the reimbursement procedure. A Q\&A session is also incorporated to address any uncertainties or concerns. After receiving their reimbursement, the annotators signed an agreement sheet to ensure its receipt.
\subsection{NEI Claim Verification Interface}
\label{append:claim validation platform}
\begin{figure*}[!tbh]
\centering
    \includegraphics[width=15cm]{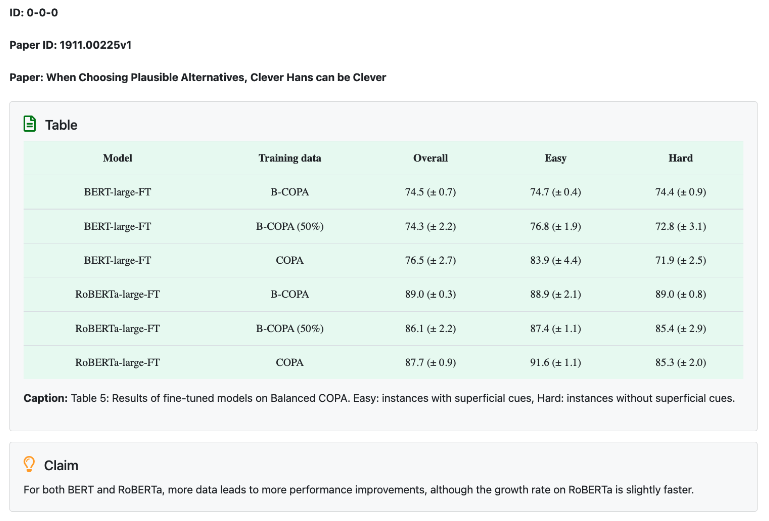}
    \caption{The user interface for the NEI claim verification task.}
    \label{fig:claim validation ui}
\end{figure*}

Figure~\ref{fig:claim validation ui} shows the user interface for the NEI claim verification task. 
\subsection{Refuted Claim Verification Interface}
\begin{figure*}[!tbh]
\centering
    \includegraphics[width=14cm]{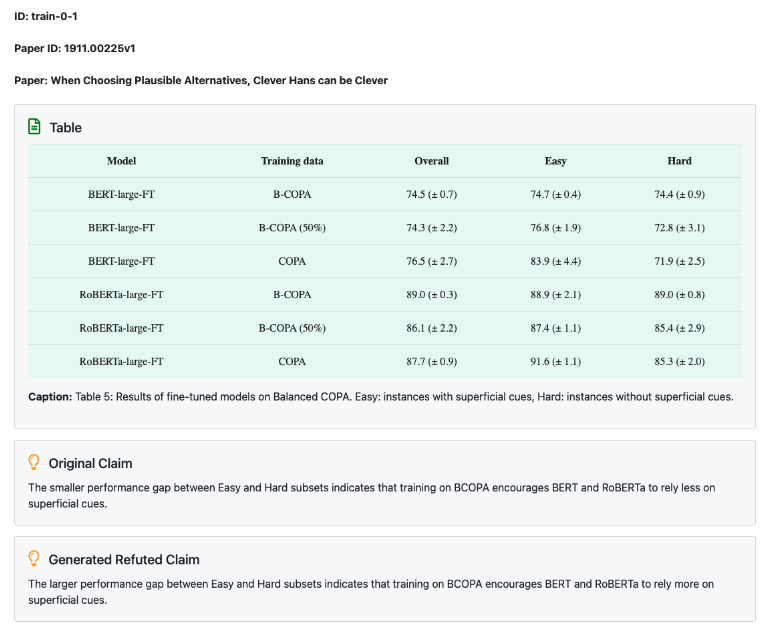}
    \caption{The user interface for the refuted claim verification task}
    \label{fig:refuted claim validation ui}
\end{figure*}

Figure~\ref{fig:refuted claim validation ui} shows the user interface for the refuted claim verification task.
\label{append: refuted claim validation interface}
\subsection{Annotation Post-Survey}
\begin{figure*}[!tbh]
    \centering
    \includegraphics[width=14cm]{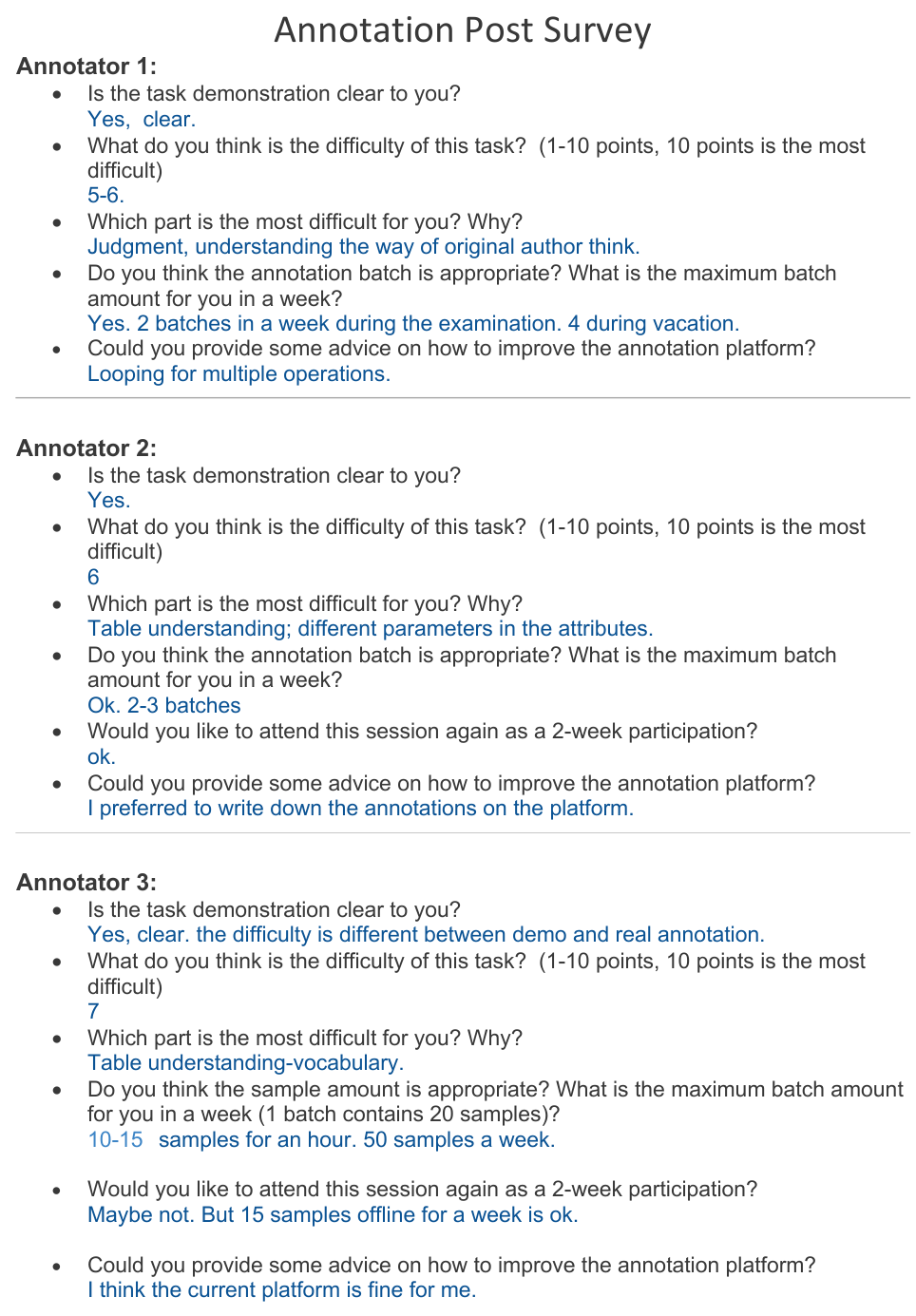}
    \caption{The examples of post-annotation survey questions and the answers of annotators.}
    \label{fig:post survey}
\end{figure*}

Figure~\ref{fig:post survey} shows the examples of post-annotation survey questions and the answers of annotators.

% \section{Reasoning Functions and Descriptions in the \dataset}
% \label{append:reasoningtab}
% Table~\ref{tab:reasoningdescriptions} shows the reasoning functions and their descriptions in the \dataset.

\section{Analysis of Refuted Reasons in the Sci-Fact dataset}
\label{append: scifact refuted reason analysis}
 Table~\ref{tab:scifact_refuted_reason} provides an analysis of the reasons for refuted claims in the Sci-Fact dataset, along with their estimated proportions. A random sample of 100 refuted claims was selected, and the results indicate that 85\% of claims were simply negated using terms like ``not'' or paraphrased based on the evidence sentences. Additionally, 6\% of the refuted claims were attributed to incorrect calculation results, while 6\% were identified as having wrong commonsense knowledge. A smaller proportion of refuted claims (3\%) were found to have incorrect open-domain knowledge.
 \begin{table}[!thb]
\centering
%\scriptsize
\resizebox{0.48\textwidth}{!}{
\begin{tabular}{lc}
\toprule
\textbf{Refuted Reasons}& \textbf{Prop. (\%)}\\\toprule
Negation (+not) and paraphrasing. & 85\\
The calculation result is wrong. & 6\\
The commonsense knowledge is wrong. & 6\\
The open-domain knowledge is wrong. & 3\\
\bottomrule
\end{tabular}
}
\caption{The refuted reasons and their estimated proportions (Prop.) in the Sci-Fact dataset. 
}
\label{tab:scifact_refuted_reason}
\vspace{-0.3cm}
\end{table}

\section{Discussions on Human-Machine Collaboration}
Our final data creation pipeline undergoes repetitive testing and revision until it reaches its current form. In our pilot annotation, we found that manual verification played the most essential role in the validation of claims marked as ``Not Enough Information(NEI)''. Initially, we planned to rely solely on LLMs for generating NEI claims. Our criteria for the NEI claim is that ``the claim should be fluent, logical, and relevant to the table. However, the claim cannot be verified as true or false solely based on the information in the table.'' However, after a careful examination of the LLM output, we found that LLM tends to generate claims that are either not logical or irrelevant to the table content. Therefore, human efforts are required to further select NEI claims that meet our criteria. Out of 900 initial NEI claims generated by LLMs, manual verification narrowed them down to only 355 claims, taking up 40\% of the original count. While it may not have served as crucial a role as filtering NEI claims, human verification also safeguarded the data quality in other annotation processes. For example, among the ``supported'' claims originally appearing in the scientific paper, human validation still identified 10 cases that were actually not supported  (\textit{e.g}., wrong number matching.)

%Claim Extraction

%Manual verification

%Refuted claim
%Post Survey

\section{Case Study for Refuted Claims}
\label{append:real refuted claim case study}
\begin{figure*}[!tbh]
    \centering
    \includegraphics[width=13cm]{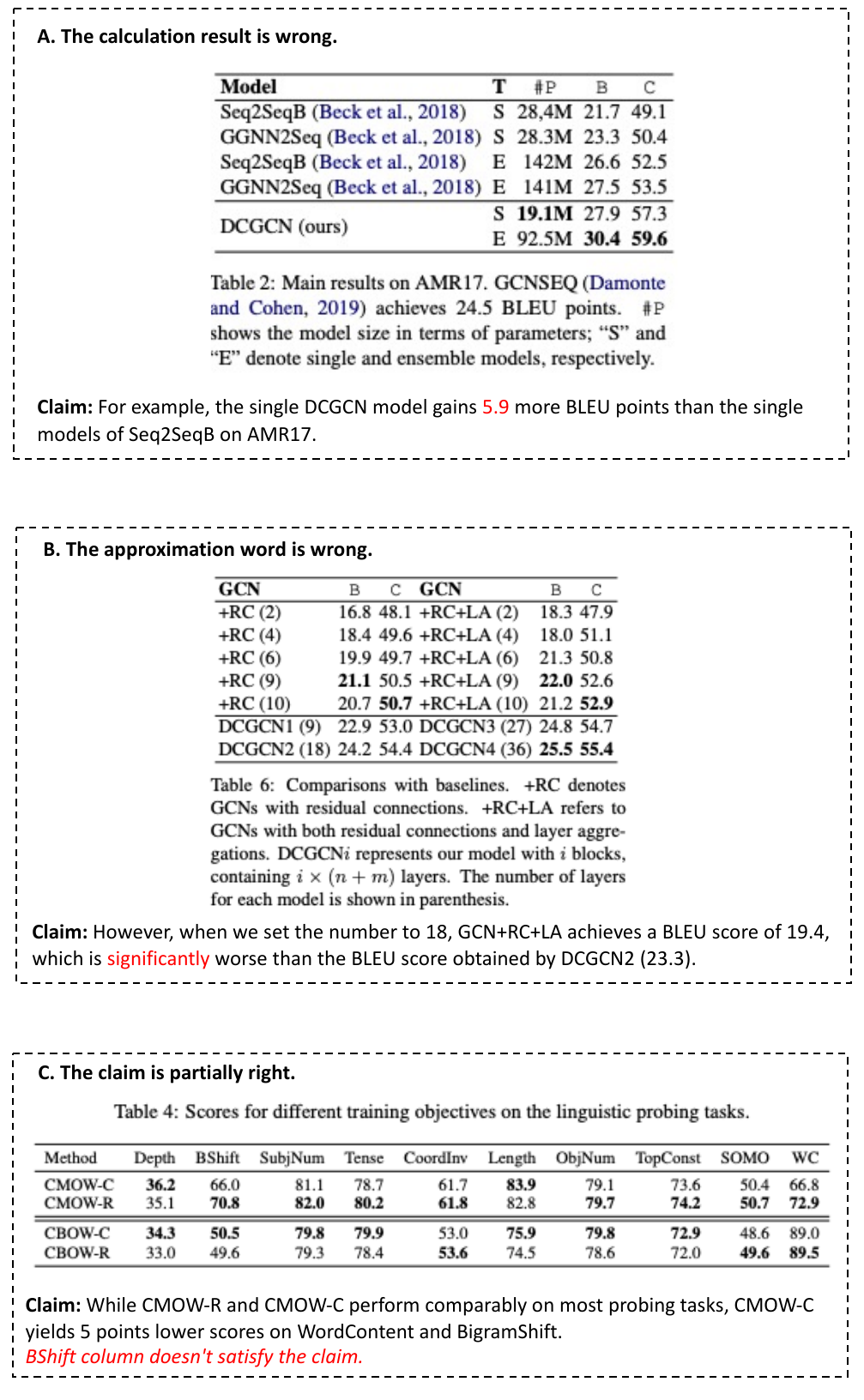}
    \caption{The refuted claims cases \textit{A} to \textit{C}. Case \textit{A} represents the calculation result is wrong. Case \textit{B} represents the approximation word is wrong. Case \textit{C} represents the claim is partially right.}
    \label{fig:refuted_cases_1}
    \vspace{-0.3cm}
    
\end{figure*}

\begin{figure*}[!tbh]
    \centering
    \includegraphics[width=13cm]{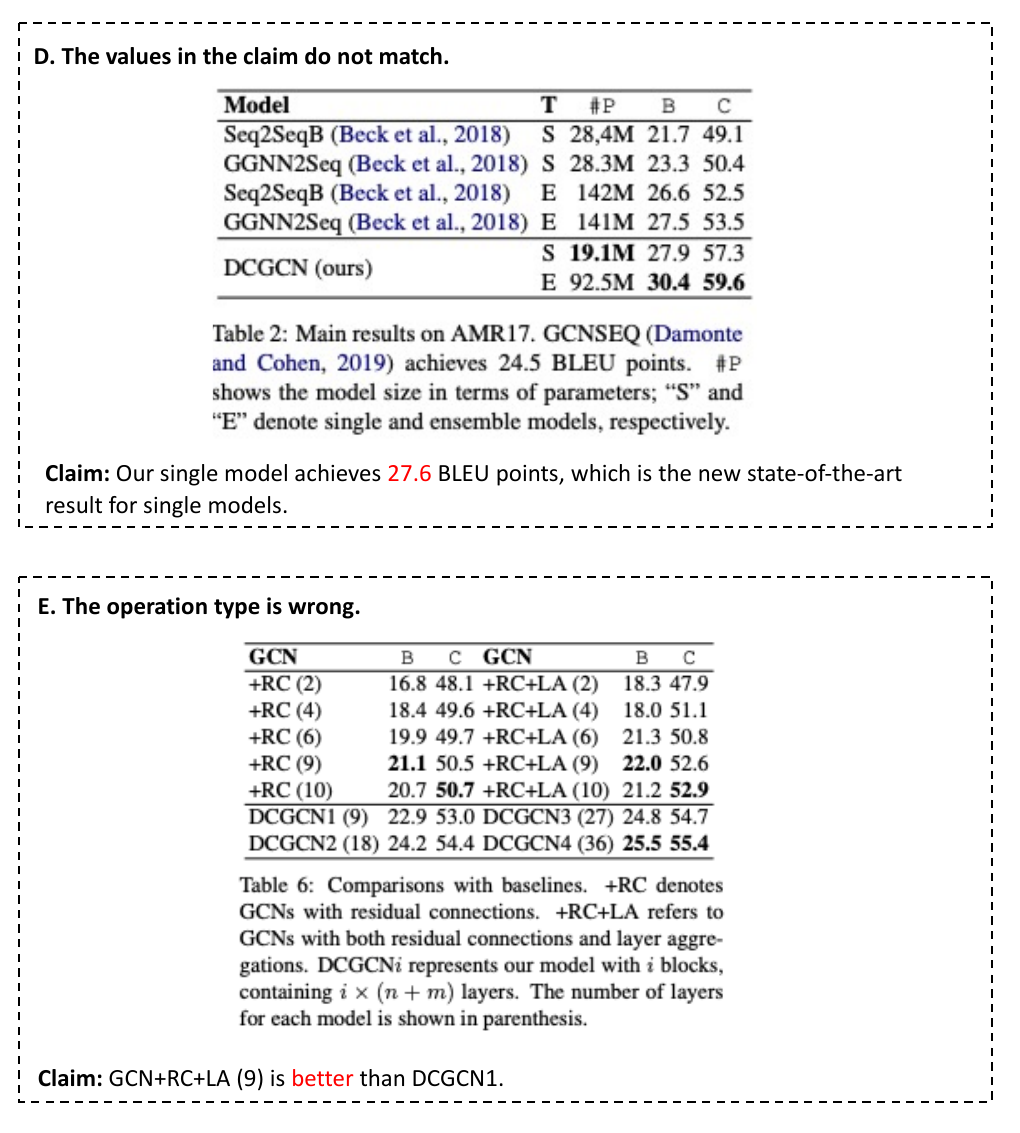}
    \caption{The refuted claims cases \textit{D} and \textit{E}. Case \textit{D} represents the values in the claim do not match. Case \textit{E} represents the operation type is wrong.}
    \label{fig:refuted_cases_2}
    \vspace{-0.3cm}
\end{figure*}

Figure~\ref{fig:refuted_cases_1} and Figure~\ref{fig:refuted_cases_2} show five examples of refuted cases. Below, we provide explanations for each of these error cases.

\paragraph{Case \textit{A} The calculation result is wrong.} It produces incorrect calculation results. The accurate result should be \texttt{27.9-21.7 = 6.2}. 

\paragraph{Case \textit{B} The approximation word is wrong.} It generates incorrect approximation words, as 19.4 is not significantly lower compared to 23.3.

\paragraph{Case \textit{C} The claim is partially right.} The claim is generally correct, with the exception of the \texttt{BShift column} which does not fulfill the claim.

\paragraph{Case \textit{D} The values in the claim do not match.} The value in the claim does not align with the corresponding value in the table. The correct value should be \texttt{27.9}.

\paragraph{Case \textit{E} The operation type is wrong.} It applies the incorrect operation type. For instance, in the case of \texttt{GCN+RC+LA (9)}, it is not accurate to claim that it is better than \texttt{DCGCN1} because \texttt{22.9 > 22.0 and 53.0 > 52.6.}

\section{Error Cases for InstructGPT}
\label{append:error cases for GPT3.5}
\begin{figure*}[!tbh]
\centering 
    \includegraphics[width=14cm]{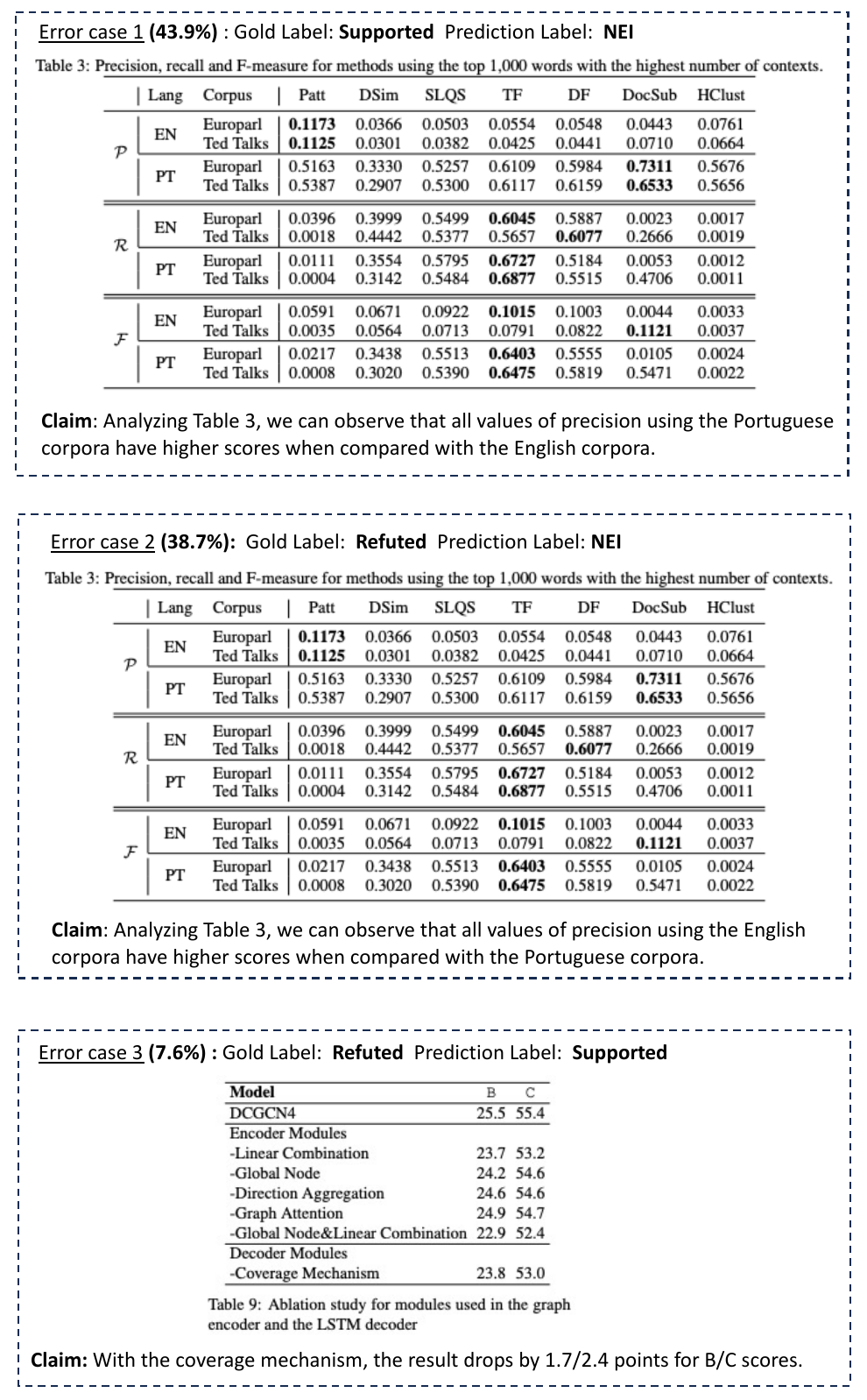}
    \caption{Error Cases 1-3 for InstructGPT in the zero-shot setting.}
    \label{fig:GPT3.5error_1}
\end{figure*}

\begin{figure*}[!tbh]
\centering 
    \includegraphics[width=14cm]{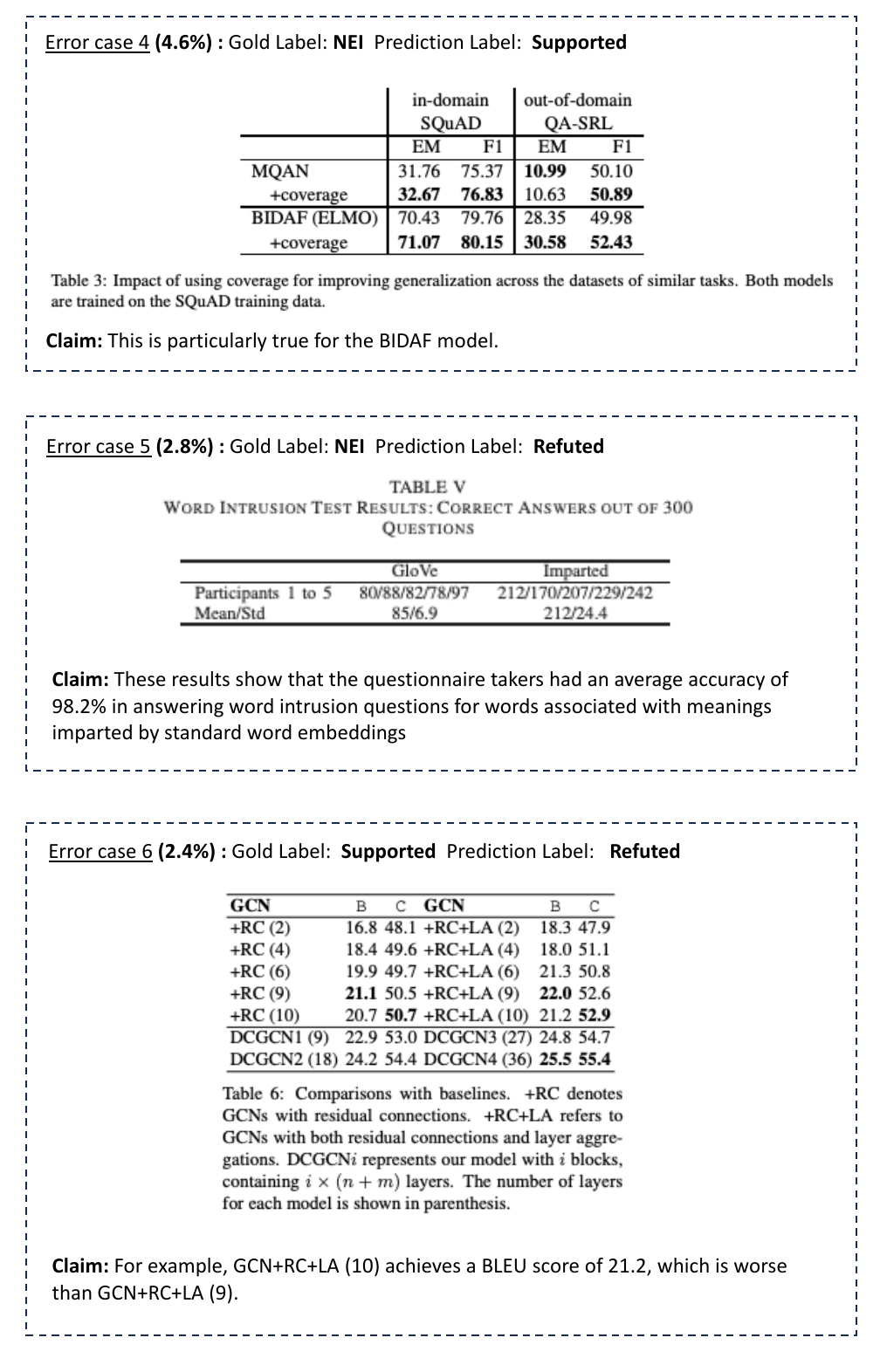}
    \caption{Error Cases 4-6 for InstructGPT in the zero-shot setting.}
    \label{fig:GPT3.5error_2}
\end{figure*}

Figure~\ref{fig:GPT3.5error_1} and Figure~\ref{fig:GPT3.5error_2} show six error examples of InstructGPT in the zero-shot setting when applied to our \dataset dataset. 
\paragraph{Error Type 1: Supported predicted as NEI.}This error type indicates a discrepancy between the gold label, which is Supported, and the predicted label, which is NEI. 
\paragraph{Error Type 2: Refuted predicted as NEI.} 
This error type indicates a discrepancy between the gold label, which is Refuted, and the predicted label, which is NEI.

\paragraph{Error Type 3: Refuted predicted as Supported.} 
This error type indicates a discrepancy between the gold label, which is Refuted, and the predicted label, which is Supported.

\paragraph{Error Type 4: NEI predicted as Supported.} 
This error type indicates a discrepancy between the gold label, which is NEI, and the predicted label, which is Supported.
\paragraph{Error Type 5: NEI predicted as Refuted.}
This error type indicates a discrepancy between the gold label, which is NEI, and the predicted label, which is Refuted.
\paragraph{Error Type 6: Supported predicted as Refuted.} 
This error type indicates a discrepancy between the gold label, which is Supported, and the predicted label, which is Refuted.

\section{Error Cases for Program-of-Thoughts}
\label{append:pot_error}
\begin{figure*}[!tbh]
\centering 
    \includegraphics[width=15cm]{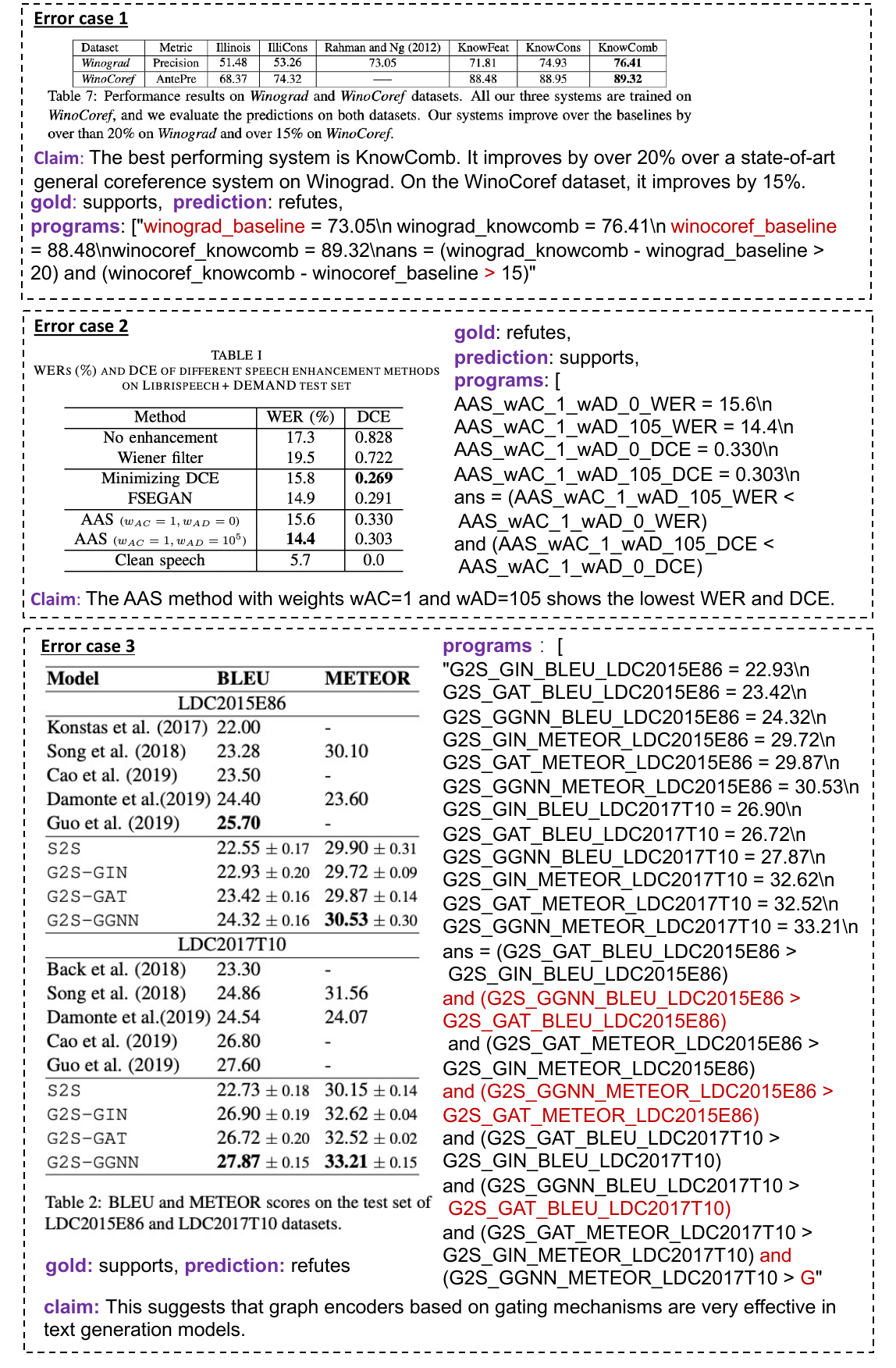}
    \caption{Error Cases 1-3 for Program-of-Thoughts. Error Case 1 exhibits incorrect entity linking (\textit{Grounding error}) and incorrect operation (\textit{Program error}). Error Case 2 exhibits incomplete entity linking (\textit{Grounding error}). Error Case 3 exhibits \textit{Program error} since it fails to generate a correct program.}
    \label{fig:PoTerror_1}
\end{figure*}

\begin{figure*}[!tbh]
\centering 
    \includegraphics[width=14cm]{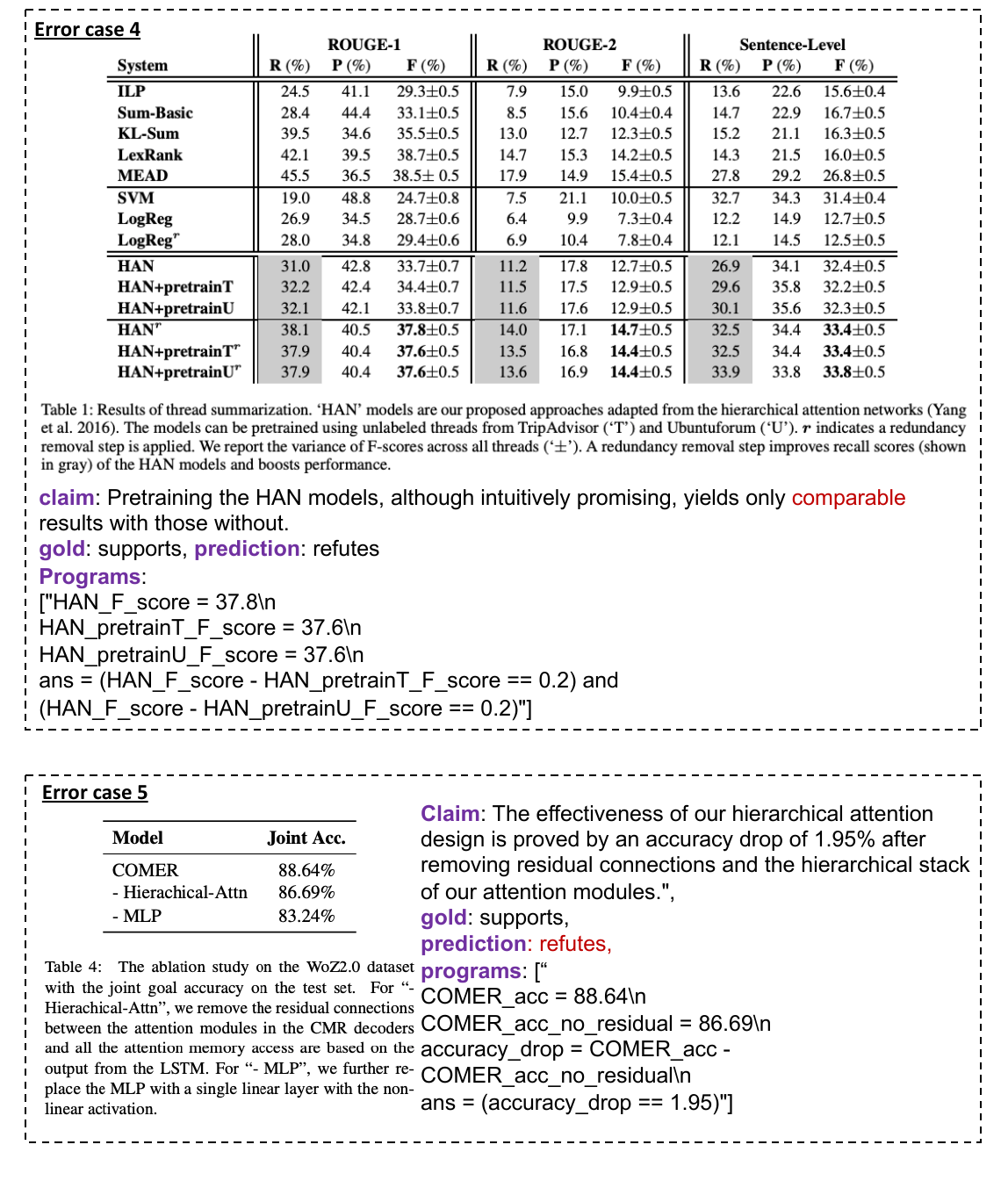}
    \caption{Error Cases 4 and 5 for Program-of-Thoughts. Error Case 4 exhibits \textit{Ambiguity error} since it fails to generate a precise program for the approximation word ``comparable''. Error Case 5 exhibits \textit{Calculation error} since it generates the correct program, but the calculation result is inaccurate due to incorrect float digits in the Python code.}
    \label{fig:PoTerror_2}
\end{figure*}

Figure~\ref{fig:PoTerror_1} and Figure~\ref{fig:PoTerror_2} show five error examples of Program-of-Thoughts when applied to our \dataset dataset. Below, we provide explanations for each of the error cases.

\paragraph{Error Case 1.} 
It exhibits incorrect entity linking (\textit{Grounding error}) and incorrect operation (\textit{Program error}).
The codes ``\texttt{winograd\_baseline = 73.06}'' and ``\texttt{winocoref\_baseline = 88.48}'' 
should be ``\texttt{IlliCons\_winograd = 53.26}'' and ``\texttt{IlliCons\_winocoref = 74.32}'' respectively. Additionally, the ``\texttt{>}'' operation should be changed to ``\texttt{>=}''.

\paragraph{Error Case 2.} It exhibits incomplete entity linking (\textit{Grounding error}). The program should also parse other baseline results, such as `\texttt{SFEGAN\_WER = 14.9}''.

\paragraph{Error Case 3.} It fails to generate a correct program (\textit{Program error}). The variables and logical functions in the programs are incorrect. For instance, ``\texttt{G2S\_GAT\_BLEU\_LDC2015E86}'' should be ``\texttt{G2S\_GIN\_BLEU\_LDC2015E86}''. The logical function ``\texttt{and}'' should be replaced with ``\texttt{or}''.

\paragraph{Error Case 4.} It fails to generate a precise program for the approximation word ``comparable'' (\textit{Ambiguity error}). Currently, the program defines ``comparable'' as ``larger than'', which is not accurate enough.

\paragraph{Error Case 5.} It generates the correct program, but the calculation result is inaccurate due to incorrect float digits in the Python code (\textit{Calculation error}). For instance, Python may output '1.9499999', which is not equal to '1.95'.

\newpage
\onecolumn
\section{Prompts}

\subsection{Zero-shot Prompts}

\lstset{
    style=mystyle,
    basicstyle=\ttfamily\scriptsize,
    backgroundcolor=\color{white},
    stringstyle=\color{black},
    keywordstyle=\color{black},
    breaklines=false,
    keepspaces=false
}
\begin{lstlisting}[language=Python]
(*@\color{codepurple}{\textbf{Table}}@*): (*@{\color{darkblue}{\textsc{<input\_table>}}@*)

(*@\color{codepurple}{\textbf{Claim}}@*): (*@{\color{darkblue}{\textsc{<input\_claim>}}@*)

Based on the information in the Table, is the above claim true?
A) the claim is true.
B) the claim is false.
C) it is impossible to tell.
\end{lstlisting}

\subsection{Few-shot Prompts}

\begin{lstlisting}[language=Python]
Read the following table and then answer a question.

(*@\color{codepurple}{\textbf{Caption}}@*): Table 5: Results of fine-tuned models on Balanced COPA. Easy: instances with superficial cues, 
Hard: instances without superficial cues.

(*@\color{codepurple}{\textbf{Table}}@*):
|| Model | Training data | Overall | Easy | Hard ||
|| BERT-large-FT | B-COPA | 74.5 ((*@$\pm$@*)0.7) | 74.7 ((*@$\pm$@*)0.4) | 74.4 ((*@$\pm$@*)0.9) ||
|| BERT-large-FT | B-COPA (50%) | 74.3 ((*@$\pm$@*)2.2) | 76.8 ((*@$\pm$@*)1.9) | 72.8 ((*@$\pm$@*)3.1) ||
|| BERT-large-FT | COPA |  76.5 ((*@$\pm$@*)2.7) |  83.9 ((*@$\pm$@*)4.4) | 71.9 ((*@$\pm$@*)2.5) ||
|| RoBERTa-large-FT | B-COPA |  89.0 ((*@$\pm$@*)0.3) | 88.9 ((*@$\pm$@*)2.1) |  89.0 ((*@$\pm$@*)0.8) ||
|| RoBERTa-large-FT | B-COPA (50%) | 86.1 ((*@$\pm$@*)2.2) | 87.4 ((*@$\pm$@*)1.1) | 85.4 ((*@$\pm$@*)2.9) ||
|| RoBERTa-large-FT | COPA | 87.7 ((*@$\pm$@*)0.9) |  91.6 ((*@$\pm$@*)1.1) | 85.3 ((*@$\pm$@*)2.0) ||

(*@\color{codepurple}{\textbf{Claim}}@*): RoBERTa-large outperforms BERT-large when fine-tuned on full and balanced COPA.

(*@\color{codepurple}{\textbf{Question}}@*): Is the above claim true or false? Please directly give the answer.

(*@\color{codepurple}{\textbf{Answer}}@*):  
The claim is true.
------
(*@\color{codepurple}{\textbf{Caption}}@*): Table 5: Results of fine-tuned models on Balanced COPA. Easy: instances with superficial cues, 
Hard: instances without superficial cues.

(*@\color{codepurple}{\textbf{Table}}@*):
|| Model | Training data | Overall | Easy | Hard ||
|| BERT-large-FT | B-COPA | 74.5 ((*@$\pm$@*)0.7) | 74.7 ((*@$\pm$@*)0.4) | 74.4 ((*@$\pm$@*)0.9) ||
|| BERT-large-FT | B-COPA (50%) | 74.3 ((*@$\pm$@*)2.2) | 76.8 ((*@$\pm$@*)1.9) | 72.8 ((*@$\pm$@*)3.1) ||
|| BERT-large-FT | COPA |  76.5 ((*@$\pm$@*)2.7) |  83.9 ((*@$\pm$@*)4.4) | 71.9 ((*@$\pm$@*)2.5) ||
|| RoBERTa-large-FT | B-COPA |  89.0 ((*@$\pm$@*)0.3) | 88.9 ((*@$\pm$@*)2.1) |  89.0 ((*@$\pm$@*)0.8) ||
|| RoBERTa-large-FT | B-COPA (50%) | 86.1 ((*@$\pm$@*)2.2) | 87.4 ((*@$\pm$@*)1.1) | 85.4 ((*@$\pm$@*)2.9) ||
|| RoBERTa-large-FT | COPA | 87.7 ((*@$\pm$@*)0.9) |  91.6 ((*@$\pm$@*)1.1) | 85.3 ((*@$\pm$@*)2.0) ||

(*@\color{codepurple}{\textbf{Claim}}@*): The difference between RoBERTa-large-FT and BERT-large-FT is 3.8 points on B-COPA, 
which is significantly smaller than the difference in COPA.

(*@\color{codepurple}{\textbf{Question}}@*): Is the above claim true or false? Please directly give the answer.

(*@\color{codepurple}{\textbf{Answer}}@*): 
The claim is false.
------
(*@\color{codepurple}{\textbf{Caption}}@*):  Table 4: The ablation study on the WoZ2.0 dataset with the joint goal accuracy on the test set. 
For``- Hierachical-Attn'', we remove the residual connections between the attention modules in the CMR 
decoders and all the attention memory access are based on the output from the LSTM. 
For``- MLP'', we further replace the MLP with a single linear layer with the non-linear activation.

(*@\color{codepurple}{\textbf{Table}}@*): 
|| Model | Joint Acc. ||
|| COMER | 88.64% ||
|| - Hierachical-Attn | 86.69% ||
|| - MLP | 83.24% ||

(*@\color{codepurple}{\textbf{Claim}}@*): [CONTINUE] The effectiveness of our hierarchical attention design is proved by an accuracy drop 
of 1.95% after removing residual connections and the hierarchical stack of our attention modules.

(*@\color{codepurple}{\textbf{Question}}@*): Is the above claim true or false? Please directly give the answer.

(*@\color{codepurple}{\textbf{Answer}}@*): 
The claim is true.
------
(*@\color{codepurple}{\textbf{Caption}}@*): Table 4: Scores for different training objectives on the linguistic probing tasks.

(*@\color{codepurple}{\textbf{Table}}@*):
|| Method | Depth| BShift| SubjNum | Tense | CoordInv | Length | ObjNum | TopConst | SOMO | WC ||
|| CMOW-C | 36.2 | 66.0 | 81.1 | 78.7 | 61.7 | 83.9 | 79.1 | 73.6 | 50.4 | 66.8 ||
|| CMOW-R | 35.1 | 70.8 | 82.0 | 80.2 | 61.8 | 82.8 | 79.7 | 74.2 | 50.7 | 72.9 ||
|| CBOW-C | 34.3 | 50.5 | 79.8 | 79.9 | 53.0 | 75.9 | 79.8 | 72.9 | 48.6 | 89.0 ||
|| CBOW-R | 33.0 | 49.6 | 79.3 | 78.4 | 53.6 | 74.5 | 78.6 | 72.0 | 49.6 | 89.5 ||

(*@\color{codepurple}{\textbf{Claim}}@*): While CMOW-R and CMOW-C perform comparably on most probing tasks, 
CMOW-C yields 5 points higher scores on WordContent and BigramShift.

(*@\color{codepurple}{\textbf{Question}}@*):  Is the above claim true or false? Please directly give the answer.

(*@\color{codepurple}{\textbf{Answer}}@*): 
The claim is false.

(*@\color{codegray}{\textbf{($\cdots$ more in-context examples here $\cdots$)}}@*)

------
(*@\color{codepurple}{\textbf{Caption}}@*): (*@{\color{darkblue}{\textsc{<input\_caption>}}@*)

(*@\color{codepurple}{\textbf{Table}}@*): (*@{\color{darkblue}{\textsc{<input\_table>}}@*)

(*@\color{codepurple}{\textbf{Claim}}@*): (*@{\color{darkblue}{\textsc{<input\_claim>}}@*)

(*@\color{codepurple}{\textbf{Question}}@*):  Is the above claim true or false? Please directly give the answer.

(*@\color{codepurple}{\textbf{Answer}}@*): 

\end{lstlisting}

\subsection{Chain-of-Thought Prompts}

\begin{lstlisting}[language=Python]
Read the following table and then answer a question.

(*@\color{codepurple}{\textbf{Caption}}@*): Table 5: Results of fine-tuned models on Balanced COPA. Easy: instances with superficial cues, 
Hard: instances without superficial cues.

(*@\color{codepurple}{\textbf{Table}}@*):
|| Model | Training data | Overall | Easy | Hard ||
|| BERT-large-FT | B-COPA | 74.5 ((*@$\pm$@*)0.7) | 74.7 ((*@$\pm$@*)0.4) | 74.4 ((*@$\pm$@*)0.9) ||
|| BERT-large-FT | B-COPA (50%) | 74.3 ((*@$\pm$@*)2.2) | 76.8 ((*@$\pm$@*)1.9) | 72.8 ((*@$\pm$@*)3.1) ||
|| BERT-large-FT | COPA |  76.5 ((*@$\pm$@*)2.7) |  83.9 ((*@$\pm$@*)4.4) | 71.9 ((*@$\pm$@*)2.5) ||
|| RoBERTa-large-FT | B-COPA |  89.0 ((*@$\pm$@*)0.3) | 88.9 ((*@$\pm$@*)2.1) |  89.0 ((*@$\pm$@*)0.8) ||
|| RoBERTa-large-FT | B-COPA (50%) | 86.1 ((*@$\pm$@*)2.2) | 87.4 ((*@$\pm$@*)1.1) | 85.4 ((*@$\pm$@*)2.9) ||
|| RoBERTa-large-FT | COPA | 87.7 ((*@$\pm$@*)0.9) |  91.6 ((*@$\pm$@*)1.1) | 85.3 ((*@$\pm$@*)2.0) ||

(*@\color{codepurple}{\textbf{Claim}}@*): RoBERTa-large outperforms BERT-large when fine-tuned on full and balanced COPA.

(*@\color{codepurple}{\textbf{Question}}@*): Is the above claim true or false? Let us think step by step.

(*@\color{codepurple}{\textbf{Reasoning}}@*):
We need to compare the performance of RoBERTa-large and BERT-large on full and balanced COPA. 
For full COPA, RoBERTa-large has an overall performance of 87.7 (table[6,2]), 
while BERT-large has an overall performance of 76.5 (table[3,2]). 
Therefore, RoBERTa-large outperforms BERT-large on full COPA. 
For balanced COPA, RoBERTa-large has an overall performance of 89.0 (table[4,2]), 
while BERT-large has an overall performance of 74.5 (table[1,2]). 
Therefore, RoBERTa-large outperforms BERT-large on balanced COPA. 
Finally, since RoBERTa-large outperforms BERT-large on both full and balanced COPA, the Claim is true.

(*@\color{codepurple}{\textbf{Answer}}@*): 
Therefore, the claim is true.
------
(*@\color{codepurple}{\textbf{Caption}}@*): Table 5: Results of fine-tuned models on Balanced COPA. Easy: instances with superficial cues, 
Hard: instances without superficial cues.

(*@\color{codepurple}{\textbf{Table}}@*):
|| Model | Training data | Overall | Easy | Hard ||
|| BERT-large-FT | B-COPA | 74.5 ((*@$\pm$@*)0.7) | 74.7 ((*@$\pm$@*)0.4) | 74.4 ((*@$\pm$@*)0.9) ||
|| BERT-large-FT | B-COPA (50%) | 74.3 ((*@$\pm$@*)2.2) | 76.8 ((*@$\pm$@*)1.9) | 72.8 ((*@$\pm$@*)3.1) ||
|| BERT-large-FT | COPA |  76.5 ((*@$\pm$@*)2.7) |  83.9 ((*@$\pm$@*)4.4) | 71.9 ((*@$\pm$@*)2.5) ||
|| RoBERTa-large-FT | B-COPA |  89.0 ((*@$\pm$@*)0.3) | 88.9 ((*@$\pm$@*)2.1) |  89.0 ((*@$\pm$@*)0.8) ||
|| RoBERTa-large-FT | B-COPA (50%) | 86.1 ((*@$\pm$@*)2.2) | 87.4 ((*@$\pm$@*)1.1) | 85.4 ((*@$\pm$@*)2.9) ||
|| RoBERTa-large-FT | COPA | 87.7 ((*@$\pm$@*)0.9) |  91.6 ((*@$\pm$@*)1.1) | 85.3 ((*@$\pm$@*)2.0) ||

(*@\color{codepurple}{\textbf{Claim}}@*): The difference between RoBERTa-large-FT and BERT-large-FT is 3.8 points on B-COPA, 
which is significantly smaller than the difference in COPA.

(*@\color{codepurple}{\textbf{Question}}@*): Is the above claim true or false? Let us think step by step.

(*@\color{codepurple}{\textbf{Reasoning}}@*):
To verify this claim, we need to compare the performance difference between RoBERTa-large-FT 
and BERT-large-FT on B-COPA and COPA. 
The difference between RoBERTa-large-FT and BERT-large-FT on B-COPA is 89.0 - 74.5 = 14.5 
(table[4,2] - table[1,2]). 
The difference between RoBERTa-large-FT and BERT-large-FT on COPA is 87.7 - 76.5 = 11.2
(table[6,2] - table[3,2]). 
The difference between RoBERTa-large-FT and BERT-large-FT on B-COPA is significantly larger 
than the difference on COPA. Therefore, the Claim is false.

(*@\color{codepurple}{\textbf{Answer}}@*):
Therefore, the claim is false.
------
(*@\color{codepurple}{\textbf{Caption}}@*):  Table 4: The ablation study on the WoZ2.0 dataset with the joint goal accuracy on the test set. 
For``- Hierachical-Attn'', we remove the residual connections between the attention modules in the CMR 
decoders and all the attention memory access are based on the output from the LSTM. 
For``- MLP'', we further replace the MLP with a single linear layer with the non-linear activation.
(*@\color{codepurple}{\textbf{Table}}@*): 
|| Model | Joint Acc. ||
|| COMER | 88.64% ||
|| - Hierachical-Attn | 86.69% ||
|| - MLP | 83.24% ||

(*@\color{codepurple}{\textbf{Claim}}@*): [CONTINUE] The effectiveness of our hierarchical attention design is proved by an accuracy drop 
of 1.95% after removing residual connections and the hierarchical stack of our attention modules.

(*@\color{codepurple}{\textbf{Question}}@*): Is the above claim true or false? Let us think step by step.

(*@\color{codepurple}{\textbf{Reasoning}}@*):
To verify this claim, we need to compare the performance of COMER with 
and without hierarchical attention design.
The performance of COMER with hierarchical attention design is 88.64% (table[2,2]). 
The performance of COMER without hierarchical attention design is 86.69% (table[3,2]). 
Therefore, the performance of COMER with hierarchical attention design is higher than 
the performance of COMER without hierarchical attention design by 88.64 - 86.69 = 1.95. 
Therefore, the Claim is true.

(*@\color{codepurple}{\textbf{Answer}}@*):
Therefore, the claim is true.
------
(*@\color{codepurple}{\textbf{Caption}}@*): Table 4: Scores for different training objectives on the linguistic probing tasks.

(*@\color{codepurple}{\textbf{Table}}@*):
|| Method | Depth| BShift| SubjNum | Tense | CoordInv | Length | ObjNum | TopConst | SOMO | WC ||
|| CMOW-C | 36.2 | 66.0 | 81.1 | 78.7 | 61.7 | 83.9 | 79.1 | 73.6 | 50.4 | 66.8 ||
|| CMOW-R | 35.1 | 70.8 | 82.0 | 80.2 | 61.8 | 82.8 | 79.7 | 74.2 | 50.7 | 72.9 ||
|| CBOW-C | 34.3 | 50.5 | 79.8 | 79.9 | 53.0 | 75.9 | 79.8 | 72.9 | 48.6 | 89.0 ||
|| CBOW-R | 33.0 | 49.6 | 79.3 | 78.4 | 53.6 | 74.5 | 78.6 | 72.0 | 49.6 | 89.5 ||

(*@\color{codepurple}{\textbf{Claim}}@*): While CMOW-R and CMOW-C perform comparably on most probing tasks, 
CMOW-C yields 5 points higher scores on WordContent and BigramShift.

(*@\color{codepurple}{\textbf{Question}}@*): Is the above claim true or false? Let us think step by step.

(*@\color{codepurple}{\textbf{Reasoning}}@*):
To verify this claim, we need to compare the performance of CMOW-R 
and CMOW-C on WordContent and BigramShift. 
The performance of CMOW-R on WordContent is 72.9 (table[2,10]). 
The performance of CMOW-C on WordContent is 66.8 (table[1,10]). 
The performance of CMOW-R on BigramShift is 70.8 (table[2,2]). 
The performance of CMOW-C on BigramShift is 66.0 (table[1,2]). 
Therefore, the performance of CMOW-C is higher than
the performance of CMOW-R on WordContent by 66.8 - 72.9 = -6.1. 
The performance of CMOW-C is lower than
the performance of CMOW-R on BigramShift by 66.0 - 70.8 = -4.8. 
Therefore, the Claim is false.

(*@\color{codepurple}{\textbf{Answer}}@*):
Therefore, the claim is false.

(*@\color{codegray}{\textbf{($\cdots$ more in-context examples here $\cdots$)}}@*)

------
(*@\color{codepurple}{\textbf{Caption}}@*): (*@{\color{darkblue}{\textsc{<input\_caption>}}@*)

(*@\color{codepurple}{\textbf{Table}}@*): (*@{\color{darkblue}{\textsc{<input\_table>}}@*)

(*@\color{codepurple}{\textbf{Claim}}@*): (*@{\color{darkblue}{\textsc{<input\_claim>}}@*)

(*@\color{codepurple}{\textbf{Question}}@*): Is the above claim true or false? Let us think step by step. 

(*@\color{codepurple}{\textbf{Reasoning}}@*):

(*@\color{codepurple}{\textbf{Answer}}@*):
\end{lstlisting}

\subsection{Program-of-Thoughts Prompts}

\begin{lstlisting}[language=Python]
Read the following table and then write Python code to answer a question: 
(please call the function equal(a, b) to check whether a and b are equal)

(*@\color{codepurple}{\textbf{Caption}}@*):  Table 4: The ablation study on the WoZ2.0 dataset with the joint goal accuracy on the test set. 
For``- Hierachical-Attn'', we remove the residual connections between the attention modules in the CMR 
decoders and all the attention memory access are based on the output from the LSTM. 
For``- MLP'', we further replace the MLP with a single linear layer with the non-linear activation.

(*@\color{codepurple}{\textbf{Table}}@*): 
|| Model | Joint Acc. ||
|| COMER | 88.64% ||
|| - Hierachical-Attn | 86.69% ||
|| - MLP | 83.24% ||

(*@\color{codepurple}{\textbf{Claim}}@*): [CONTINUE] The effectiveness of our hierarchical attention design is proved by 
an accuracy drop of 1.95% after removing residual connections 
and the hierarchical stack of our attention modules.

(*@\color{codepurple}{\textbf{Question}}@*): Based on the information in the table, is the above claim true or false?

# Python Code, return ans
COMER_acc = 88.64
COMER_acc_no_residual = 86.69
accuracy_drop = COMER_acc - COMER_acc_no_residual
ans = equal(accuracy_drop, 1.95)
------
Read the following table and then write Python code to answer a question: 
(please call the function equal(a, b) to check whether a and b are equal)

(*@\color{codepurple}{\textbf{Caption}}@*): Table 3: Ablation study of capsule net and word-level attention on Wikidata dataset.

(*@\color{codepurple}{\textbf{Table}}@*):
|| Recall | 0.1 | 0.2 | 0.3 | AUC ||
|| -Word-ATT | 0.648 | 0.515 | 0.395 | 0.389 ||
|| -Capsule | 0.635 | 0.507 | 0.413 | 0.386 ||
|| Our Model | 0.650 | 0.519 | 0.422 | 0.405 ||

(*@\color{codepurple}{\textbf{Claim}}@*): According to the table, the drop of precision demonstrates 
that the word-level attention is quite useful.

(*@\color{codepurple}{\textbf{Question}}@*): Based on the information in the table, is the above claim true or false?

# Python Code, return ans
our_model_recalls = [0.650, 0.519, 0.422, 0.405]
without_word_att_recalls = [0.648, 0.515, 0.395, 0.389]
ans = True
for i in range(4):
    if our_model_recalls[i] < without_word_att_recalls[i]:
        ans = False
        break
------
Read the following table and then write Python code to answer a question: 
(please call the function equal(a, b) to check whether a and b are equal)

(*@\color{codepurple}{\textbf{Caption}}@*): Table 4: Scores for different training objectives on the linguistic probing tasks.

(*@\color{codepurple}{\textbf{Table}}@*):
|| Method | Depth| BShift| SubjNum | Tense | CoordInv | Length | ObjNum | TopConst | SOMO | WC ||
|| CMOW-C | 36.2 | 66.0 | 81.1 | 78.7 | 61.7 | 83.9 | 79.1 | 73.6 | 50.4 | 66.8 ||
|| CMOW-R | 35.1 | 70.8 | 82.0 | 80.2 | 61.8 | 82.8 | 79.7 | 74.2 | 50.7 | 72.9 ||
|| CBOW-C | 34.3 | 50.5 | 79.8 | 79.9 | 53.0 | 75.9 | 79.8 | 72.9 | 48.6 | 89.0 ||
|| CBOW-R | 33.0 | 49.6 | 79.3 | 78.4 | 53.6 | 74.5 | 78.6 | 72.0 | 49.6 | 89.5 ||

(*@\color{codepurple}{\textbf{Claim}}@*): While CMOW-R and CMOW-C perform comparably on most probing tasks, 
CMOW-C yields 5 points higher scores on WordContent and BigramShift.

(*@\color{codepurple}{\textbf{Question}}@*): Based on the information in the table, is the above claim true or false?

# Python Code, return ans
CMOW_C_score_on_WC = 66.8
CMOW_C_score_on_BShift = 66.0
CMOW_R_score_on_WC = 72.9
CMOW_R_score_on_BShift = 70.8
ans = equal(CMOW_C_score_on_WC - CMOW_R_score_on_WC, 5) 
and equal(CMOW_C_score_on_BShift - CMOW_R_score_on_BShift, 5)

(*@\color{codegray}{\textbf{($\cdots$ more in-context examples here $\cdots$)}}@*)

------
Read the following table and then write Python code to answer a question: 
(please call the function equal(a, b) to check whether a and b are equal)

(*@\color{codepurple}{\textbf{Caption}}@*): (*@{\color{darkblue}{\textsc{<input\_caption>}}@*)

(*@\color{codepurple}{\textbf{Table}}@*): (*@{\color{darkblue}{\textsc{<input\_table>}}@*)

(*@\color{codepurple}{\textbf{Claim}}@*): (*@{\color{darkblue}{\textsc{<input\_claim>}}@*)

(*@\color{codepurple}{\textbf{Question}}@*): Based on the information in the table, is the above claim true or false?

# Python Code, return ans
\end{lstlisting}

%Case study for refuted claims
%Case for InstructGPT

%Case for Pot


\end{document}